\documentclass[11pt]{article}

\usepackage[final]{acl}
\usepackage{times}
\usepackage{latexsym}
\usepackage[T1]{fontenc}
\usepackage[utf8]{inputenc}
\usepackage{microtype}
\usepackage{inconsolata}
\usepackage{graphicx}
\usepackage{booktabs}
\usepackage{multirow}
\usepackage{xspace}
\usepackage{float}
\usepackage{caption}
\usepackage{array}
\usepackage{amsmath}
\usepackage{amssymb}
\usepackage{xcolor}
\usepackage{enumitem}
\usepackage{colortbl}
\usepackage{listings}
\usepackage[most]{tcolorbox}
\definecolor{promptcolor}{RGB}{237,242,236}
\definecolor{promptcolorheader}{RGB}{199,216,196}
\definecolor{datacolor}{RGB}{235,241,249}
\definecolor{datacolorheader}{RGB}{194,211,233}
\lstdefinestyle{prompt}{
  basicstyle=\ttfamily\scriptsize,
  breaklines=true,
  breakatwhitespace=false,
  frame=none,
  backgroundcolor=\color{promptcolor!35},
  xleftmargin=8pt,
  xrightmargin=8pt,
  aboveskip=6pt,
  belowskip=4pt,
  keepspaces=true,
  showstringspaces=false,
}
\newtcolorbox{promptbox}[2][]{
  enhanced,
  colback=promptcolor!35,
  colframe=promptcolorheader,
  boxrule=0.5pt,
  arc=1pt,
  left=0.5em,
  right=0.5em,
  top=0.4em,
  bottom=0.4em,
  title={#2},
  fonttitle=\bfseries\small,
  colbacktitle=promptcolorheader,
  coltitle=black,
  #1
}
\newtcolorbox{databox}[2][]{
  enhanced,
  colback=datacolor!55,
  colframe=datacolorheader,
  boxrule=0.5pt,
  arc=1pt,
  left=0.5em,
  right=0.5em,
  top=0.4em,
  bottom=0.4em,
  title={#2},
  fonttitle=\bfseries\small,
  colbacktitle=datacolorheader,
  coltitle=black,
  #1
}

\newcommand{\ours}{\textsc{SPARK}}
\newcommand{\oursdata}{Spark-234K}
\newcommand{\oursfull}{\textbf{\ours{}} (Scientific Paper Abstracted Reasoning sKeleton)}
\newcommand{\corpus}{\textsc{Sci-Base}}
\def\huggingface{\raisebox{-1.5pt}{\includegraphics[height=1.05em]{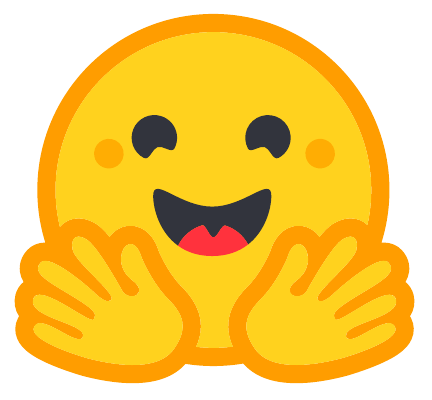}}}
\newcommand{\hflink}{https://huggingface.co/datasets/OpenDataArena/Spark-234K}

\title{
    \begin{center}
        \raisebox{-0.8ex}{\includegraphics[height=1.5em]{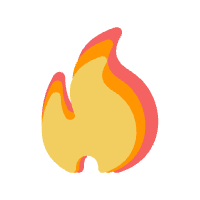}}\ours{}: Skeleton-Guided Reasoning Synthesis from \\ Large-Scale Scientific Literature
    \end{center}
}

\author{
   Yu Li\textsuperscript{1,2},
   Wei Li\textsuperscript{1,3},
   Xin Gao\textsuperscript{1},
   Mengyuan Sun\textsuperscript{4},
   {\bf  Xiaoyang Wang\textsuperscript{1},}
   {\bf Qizhi Pei\textsuperscript{5},}
   {\bf Lijun Wu\textsuperscript{1}}\thanks{Corresponding author.} \\
   \textsuperscript{1}Shanghai AI Laboratory
   \textsuperscript{2}University of Science and Technology of China  \\
    \textsuperscript{3}East China Normal University
    \textsuperscript{4}Peking University
    \textsuperscript{5}Renmin University of China \\
    \texttt{liyu1@pjlab.org.cn},
    \texttt{lijun\_wu@outlook.com} \\
    \textbf{\huggingface} \xspace \href{\hflink} {\texttt{https://huggingface.co/datasets/OpenDataArena/Spark-234K}} \\
}

\begin{document}
\maketitle

\begin{abstract}
%==============================================================================
Scientific reasoning remains challenging for open-source models, largely due to the lack of high-quality scientific reasoning data. Existing datasets are often dominated by factual recall or formulaic problem solving, with limited emphasis on mechanism understanding, evidence-grounded reasoning, and hypothesis evaluation. 
To address this, we introduce \oursfull{}, a paper-oriented synthesis framework built on \corpus{}, a large-scale corpus of research papers spanning 10 scientific disciplines. Instead of directly converting papers into question-answer pairs, \ours{} treats the \emph{claim-evidence-derivation} structure of a paper as the fundamental unit of reasoning synthesis. Specifically, \ours{} (1) distills each paper into a compact reasoning skeleton capturing its central claims and supporting evidence, enabling self-contained question generation, and (2) synthesizes reasoning tasks from four scientific perspectives: \emph{mechanistic reasoning, hypothesis falsification, quantitative derivation, and boundary calibration}. A final consistency verification stage further removes unsupported or contradictory outputs.
Using this framework, we construct \oursdata{}, a scientific reasoning dataset with substantially higher difficulty and diversity than existing resources. Experiments show that \oursdata{} consistently outperforms existing scientific reasoning datasets while achieving stronger performance with significantly fewer training samples.
\end{abstract}

%==============================================================================
% TEASER / MAIN-RESULTS FIGURE (placeholder -- fill in your actual figure)
%==============================================================================
\begin{figure}[t]
    \centering
    \includegraphics[width=\linewidth]{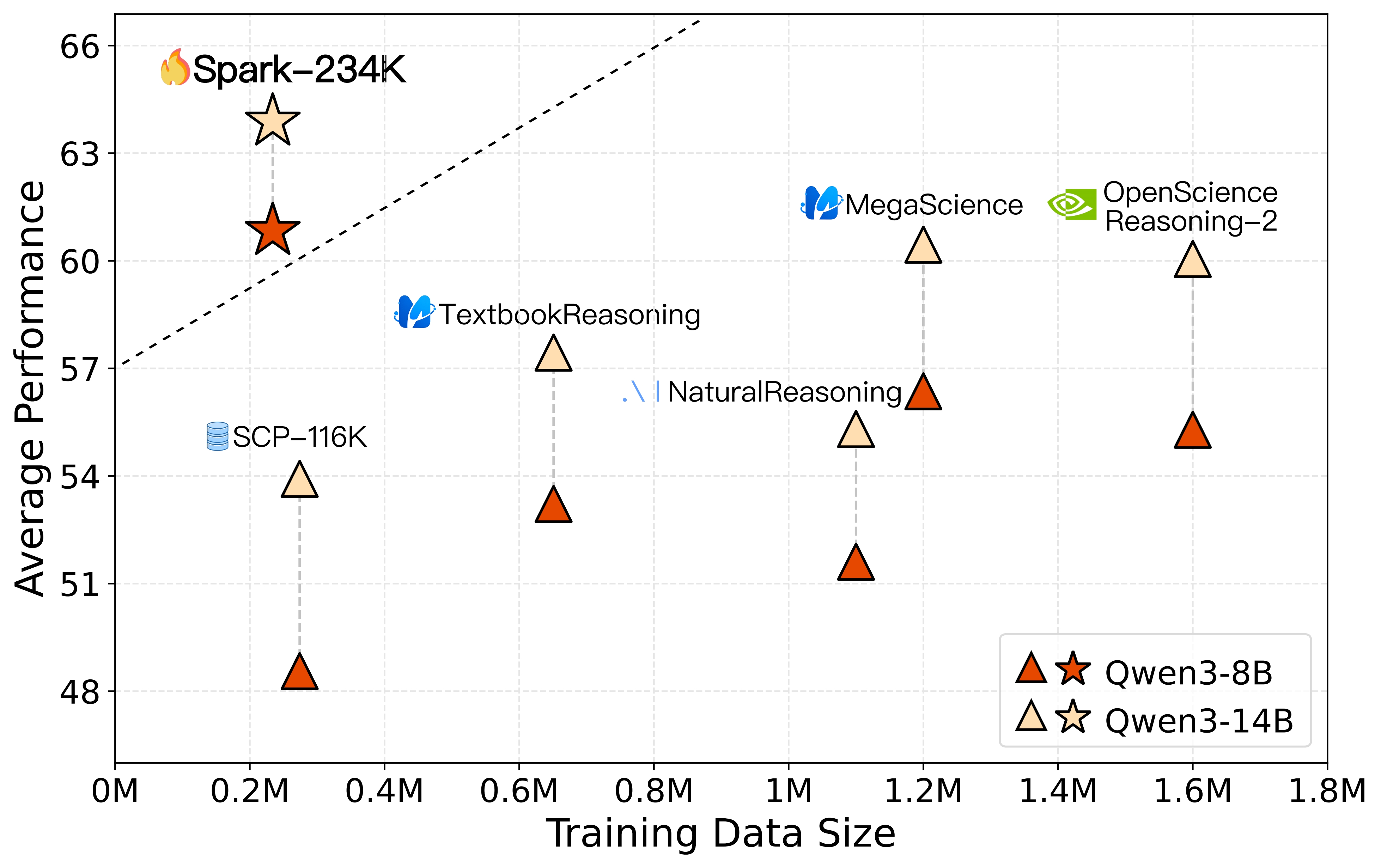}
    \caption{Comparison of the Qwen3-8B base and Qwen3-14B base models trained on baseline datasets versus \oursdata{} (benchmark scores in Table~\ref{tab:main_results}).}
    \label{fig:output}
\end{figure}

%==============================================================================
\section{Introduction}
\label{sec:intro}
Recent Large Language Models (LLMs) have achieved strong performance on reasoning tasks such as mathematics and coding~\citep{li2025domainhelpothersdatacentric,cai2025opendataarena,pei2025mathfusion}, but scientific reasoning remains significantly more challenging~\citep{li2026tracing}. A central limitation is the lack of high-quality scientific reasoning data. Existing scientific corpora are often constructed from textbooks, exam problems, or web resources~\citep{lu2025scp,fan2025megascience}, and therefore mainly emphasize factual recall, standard derivations, or routine calculations while underrepresenting the evidence-based and mechanism-driven reasoning processes that characterize real scientific discovery.

Research papers provide a natural source of scientific reasoning supervision because they contain expert-validated claims, supporting evidence, derivations, assumptions, and experimental analysis at scale~\citep{taylor2022galactica}. Recent work has also explored leveraging scientific literature for reasoning-oriented data construction and scientific question-answering~\citep{noorbakhsh2026savaal,dong-etal-2024-mc,liu-etal-2024-lost}. However, synthesizing high-quality reasoning data from papers remains difficult. Scientific papers are long, densely structured documents whose reasoning processes are often distributed across multiple sections, figures, tables, and experimental discussions. As a result, naively generating reasoning data from papers frequently produces samples that are not self-contained~\citep{choi-etal-2021-decontextualization,dasigi-etal-2021-dataset,jain-garimella-2026-knowing}, omit critical assumptions or evidence, or fail to preserve the underlying claim-evidence reasoning structure (examples are shown in Figure~\ref{fig:self_containment_bad_cases}).

We introduce \oursfull{}, a paper-oriented synthesis framework for scientific reasoning data construction. Instead of directly generating reasoning data from sections, chunks, or entire documents, \ours{} treats a paper's claim-evidence-derivation chain as the fundamental unit of synthesis.

Given a scientific paper, \ours{} first extracts its central claims together with the supporting evidence, assumptions, quantitative relations, and boundary conditions, and organizes them into a compact reasoning skeleton. This reasoning skeleton rewrites the core scientific argument into a self-contained form, reducing long-context interference~\citep{liu-etal-2024-lost} while preserving the underlying reasoning process and mitigating the lack of self-containment commonly observed in document-grounded generation~\citep{choi-etal-2021-decontextualization,dasigi-etal-2021-dataset,jain-garimella-2026-knowing}.

Based on the reasoning skeleton, \ours{} synthesizes scientific reasoning data from four perspectives: (1) \emph{mechanistic reasoning}, which explains why an observed phenomenon occurs~\citep{machamer2000thinking}; (2) \emph{hypothesis falsification}, which distinguishes competing explanations~\citep{pearl2009causality}; (3) \emph{quantitative derivation}, which emphasizes modeling and derivation instead of formula substitution~\citep{sun2024scieval}; and (4) \emph{boundary calibration}, which examines the assumptions and validity limits under which conclusions hold. A final consistency verification stage further removes unsupported or contradictory outputs.

Through the above framework and using 370K frontier scientific papers from the open \corpus{}~\cite{scibase2026} corpus, we construct \oursdata{}, a high-quality scientific reasoning dataset containing 234K synthesized instances. Despite its compact scale, this dataset exhibits strong data efficiency and overall performance (Figure~\ref{fig:output}). Notably, it outperforms massive baseline datasets such as MegaScience and OpenScienceReasoning-2~\citep{hf_opensciencereasoning2}, despite these collections being more than five and nearly seven times its size, respectively. Across distinct base model families and parameter scales, our data surpasses the runner-up datasets on multiple benchmarks by substantial average margins of $+3.10$, $+4.47$, and $+3.43$ points.

\begin{figure*}[t]
  \centering
  \small
  \setlength{\fboxsep}{6pt}
  \includegraphics[width=1.0\linewidth]{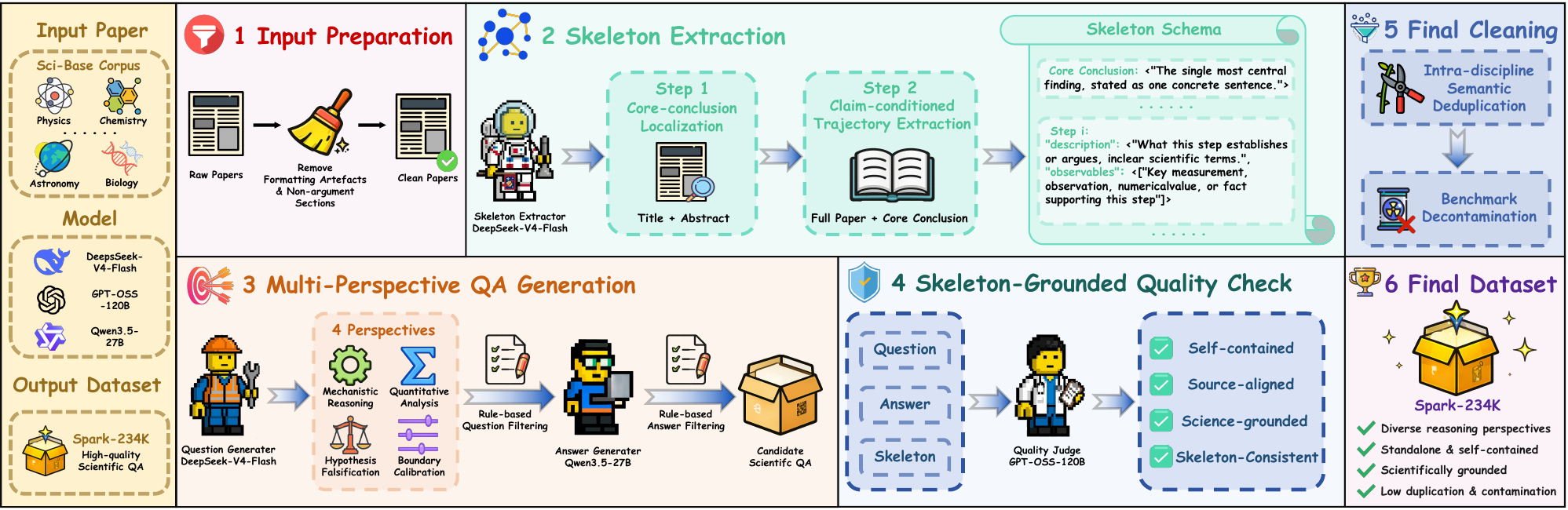}
  \caption{The \ours{} data synthesis pipeline. A frontier paper is first condensed into a \emph{reasoning skeleton}. We then generate deep scientific QA pairs via four targeted perspectives, followed by a lightweight consistency check to ensure data quality and self-containment.}
  \label{fig:pipeline}
\end{figure*}

%==============================================================================
\section{Paper Collection and Preparation}
\label{sec:method:data}

\paragraph{Source corpus.}
\ours{} is built on \corpus{}~\citep{scibase2026}, an open scientific literature corpus comprising $\sim$3.36M papers across ten disciplines. Each paper is accompanied by metadata and parsed with MinerU2.5~\citep{niu2025mineru2} into an ordered sequence of paragraphs, tables, captions, and equations, with \LaTeX{} preserved where available. This structured representation facilitates paper-level argument reconstruction while retaining the quantitative information needed for scientific reasoning.

\paragraph{Quality filtering and balanced sampling.}
To prepare inputs for the pipeline, we apply strict filtering. We restrict the timeframe to January 2024--March 2026 and discard non-English, incompletely parsed, or review articles lacking a clear trajectory of claims and evidence. Crucially, we prioritize papers rich in concrete observables, such as equations, statistical results, tables, and experimental comparisons. Finally, to mitigate the raw corpus's heavy skew toward medicine and life sciences (Table~\ref{tab:corpus_stats}), we apply discipline-balanced sampling across the ten major scientific disciplines it spans. The resulting seed pool comprises roughly 370K argument-structured frontier papers.

\section{Dataset Construction}
\label{sec:method}

Figure~\ref{fig:pipeline} illustrates the \ours{} pipeline, which converts a screened paper into standalone training instances. The key design choice is to treat a paper as an argument rather than a flat document: \ours{} first identifies the core conclusion, reconstructs the evidence trajectory that supports it, and stores this trajectory as a compact \emph{reasoning skeleton}. Questions are then generated from the skeleton under several scientific reasoning perspectives and filtered against the same skeleton for self-containment and source consistency.

\subsection{Skeleton Extraction}
\label{sec:method:skeleton}
Skeleton extraction makes a long paper usable for supervision without flattening it into sections or fragmenting it into independent chunks. Directly prompting over the full paper risks \emph{context diffusion}: the central scientific argument is diluted by peripheral details, making generation prone to shallow or unfocused questions. Instead of preserving the paper's original layout, skeleton extraction recovers the claim-centered argument: what conclusion the paper advances, what evidence supports it, and under which assumptions or boundary conditions the conclusion holds.
 
\paragraph{Input normalization.}
We first convert the parsed structure into a linearized text representation. This preserves the scientific content vital for reasoning, including equations, table text, captions, and numerical results, while stripping away formatting artifacts (such as headers, footers, and affiliations) and non-argument sections (such as references and acknowledgments).

\paragraph{Two-step extraction.}
Skeleton extraction recovers the core argument rather than the original section order. First, \emph{core conclusion localization} identifies the central claim from the title and abstract, which typically state the research question, main finding, and scope while avoiding distracting body content. Second, \emph{trajectory extraction} reads the normalized full paper conditioned on this conclusion, reconstructing the supporting argument as ordered steps. Each step contains a description of its logical role and a set of observables, such as measurements, comparisons, numerical values, formulas, or stated boundary conditions.

\begin{figure}[t]
  \centering
  \small
  \setlength{\fboxsep}{6pt}
  \includegraphics[width=1.0\linewidth]{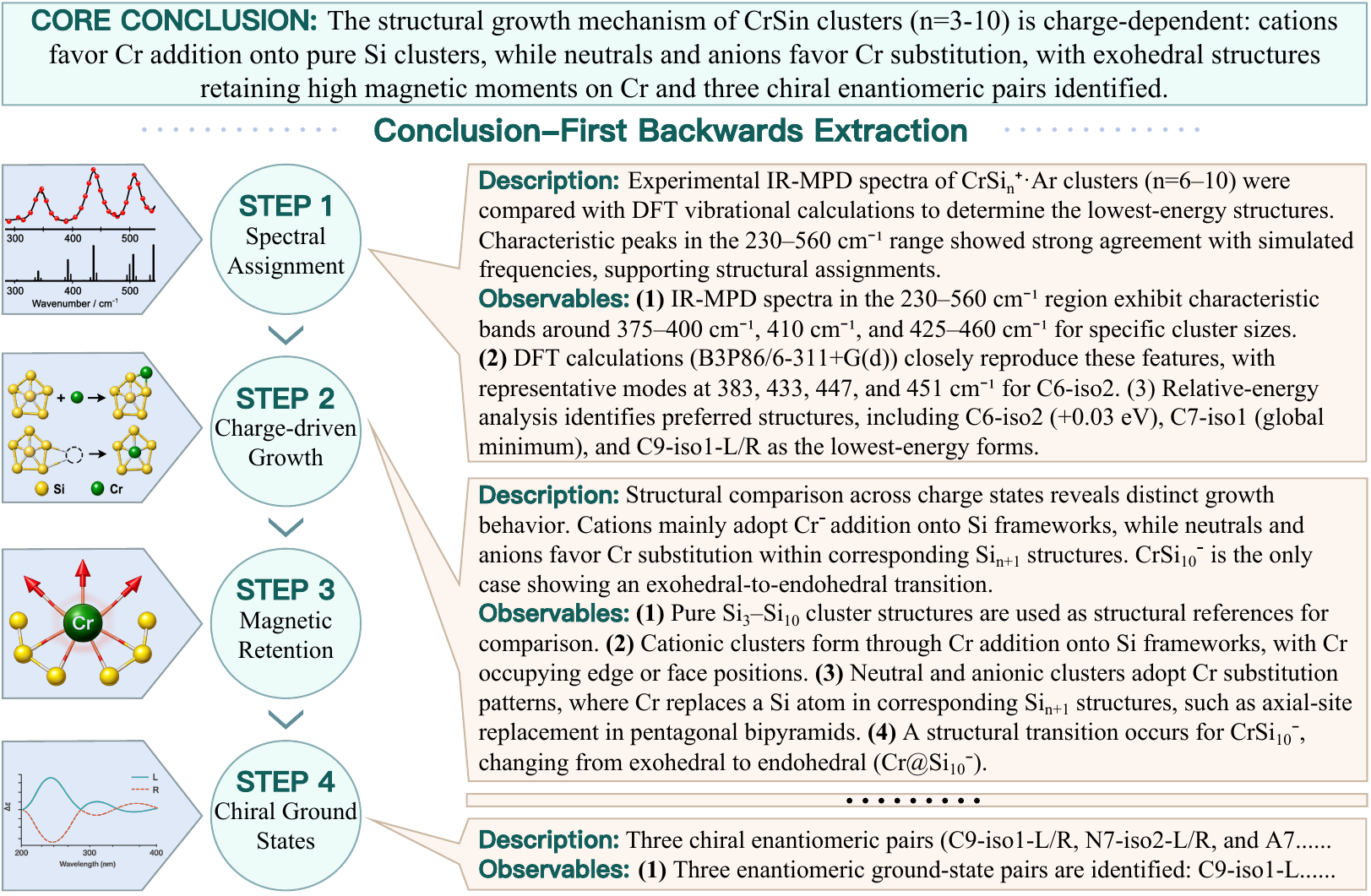}
  \caption{Example skeleton extracted by \ours{} from a real paper, condensing the source into a core conclusion and ordered evidence steps with concrete observables.}
  \label{fig:skeleton_example}
\end{figure}

The resulting skeleton comprises the core conclusion and an ordered list of steps. As Figure~\ref{fig:skeleton_example} illustrates, it is not a section-level summary~\citep{edge2025localglobalgraphrag} but a compact evidence trajectory organized entirely around the paper's central claim.

\subsection{Multi-Perspective Question Generation}
\label{sec:method:generation}
 
The next stage generates questions that probe scientific reasoning rather than surface facts. The skeleton exposes the paper's claim, evidence, inferential links, and boundary conditions, which must be connected to understand the argument~\citep{javaji-etal-2025-ai}. We therefore generate questions from four complementary perspectives.

\textbf{Mechanistic reasoning} turns a reported condition--effect relation into a causal explanation problem, asking why a structure, intervention, or regime produces the phenomenon~\citep{machamer2000thinking}. \textbf{Hypothesis falsification} uses contrasts, ablations, or ruled-out alternatives to ask which explanatory model is supported by the evidence~\citep{pearl2009causality}. \textbf{Quantitative derivation} converts numerical observables, formulas, or scaling relations into modeling-first problems whose difficulty lies in selecting and linking principles, rather than substituting values~\citep{sun2024scieval}. \textbf{Boundary calibration} asks where a conclusion ceases to be justified, focusing on the assumptions, regimes, confounders, or approximations that delimit the claim.

Together, these perspectives ask why a result holds, which explanation survives the evidence, how quantities are related, and when the claim breaks down. This shifts generation away from fact retrieval and plug-and-chug computation toward mechanism-level reasoning. To maintain precision, \ours{} follows a conservative yielding strategy: for each paper, it uses only the perspectives recommended during skeleton extraction, generates at most one question per perspective, and skips any perspective unsupported by the evidence.

\paragraph{Filtering.}
Generated questions undergo a lightweight, rule-based filtering stage to detect malformed formats and source-leakage phrases (detailed in Appendix~\ref{sec:appendix:leakage}). The central constraint is self-containment: a question depending on unstated figures, tables, or source context is discarded. Finally, questions are audited for information sufficiency and intellectual depth, with thresholds adjusted to prevent the over-correction of valid reasoning trajectories.
\subsection{Answer Generation}
\label{sec:method:answer}
An answer model responds to each accepted question without access to the skeleton or source text. This setup mirrors the downstream fine-tuning environment and empirically verifies that the question is self-contained. We then apply heuristic filters to discard answers that are malformed, excessively short or long, or dominated by repeated $n$-grams.

\subsection{Skeleton-Grounded Quality Check}
\label{sec:method:quality}

While rule-based checks catch formatting issues, they cannot assess scientific grounding. We therefore deploy an LLM judge to evaluate each $\langle$\,question, answer, skeleton\,$\rangle$ triple against four criteria. For the \textbf{question}, it verifies: (i) self-containment (supplying necessary context independent of the paper), and (ii) consistency with the skeleton. For the \textbf{answer}, it checks if the reasoning is (iii) coherent and (iv) logically supported by the evidence trajectory. QA pairs failing any criterion are discarded. 

The self-containment check acts primarily as a safeguard, as skeleton extraction prevents most dependencies at the source (evidenced by the low discard rate in Table~\ref{tab:pipeline_yield}). Crucially, by grounding the answer evaluation directly in the extracted skeleton, we bypass the massive majority voting and iterative rewriting typically required in traditional math and reasoning domains~\citep{tong2024dartmath}, keeping the pipeline computationally lightweight.

\subsection{Final Data Cleaning}
\label{sec:method:cleaning}
Finally, we apply deduplication and decontamination procedures. We perform semantic deduplication within each discipline using Qwen3-Embedding-8B embeddings~\citep{zhang2025qwen3}. To prevent benchmark leakage, we enforce 13-gram exact matching against all downstream evaluation sets, discarding any overlapping records.

%==============================================================================
\section{Dataset Statistics and Analysis}
\label{sec:stats}
%==============================================================================

By applying the proposed synthesis pipeline to the 370K filtered seed papers, we construct \oursdata{}, a high-quality instruction tuning dataset dedicated to deep scientific reasoning.

\subsection{Dataset Statistics}
\label{sec:dataset_statistics}

\begin{figure}[t]
    \centering
    \includegraphics[width=\columnwidth]{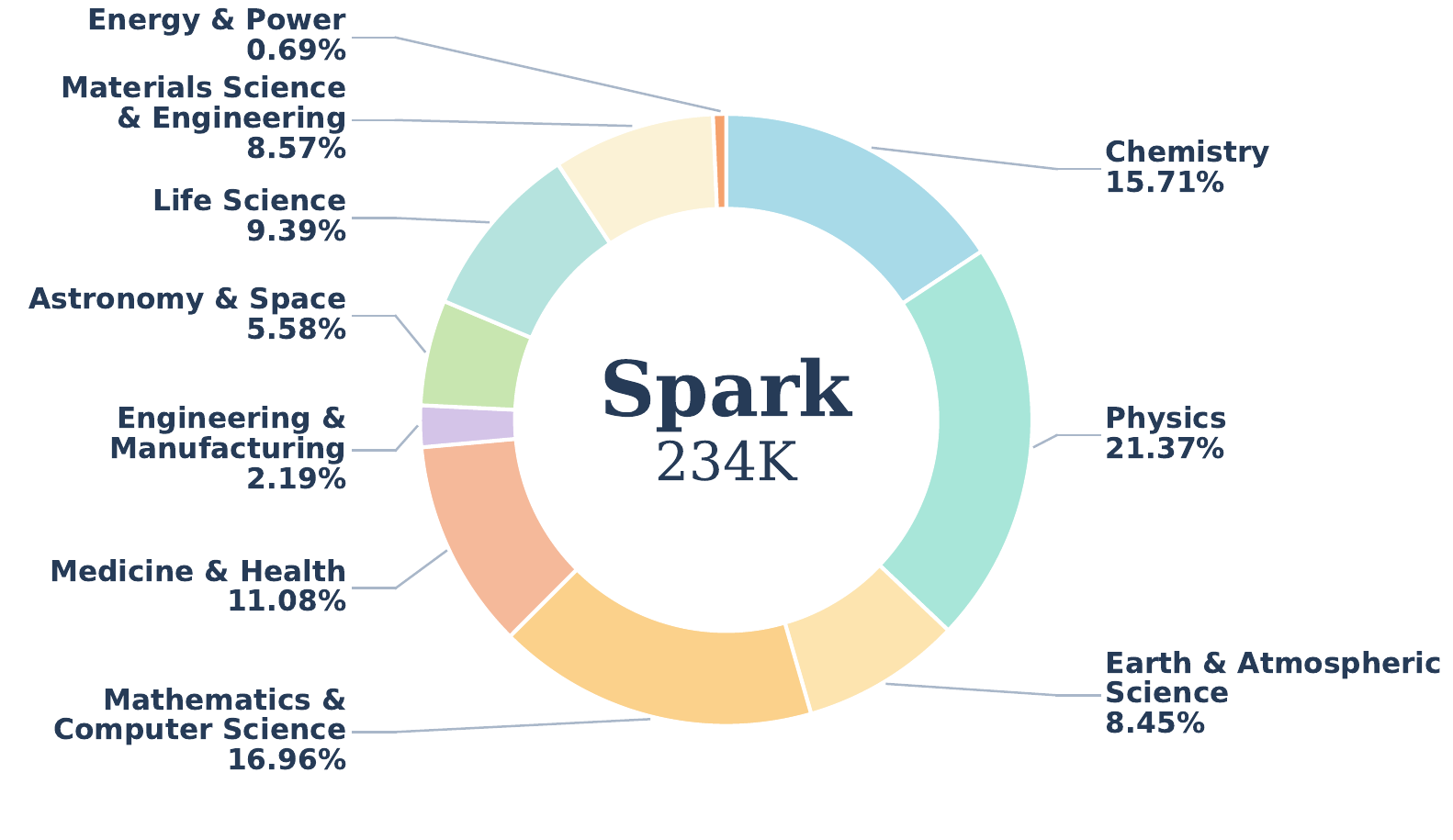}
    \caption{Discipline distribution of \oursdata{}.}
    \label{fig:Disciplinary}
\end{figure}

Figure~\ref{fig:Disciplinary} illustrates the disciplinary distribution of \oursdata{}. Physics (21.37\%), Mathematics \& Computer Science (16.96\%), and Chemistry (15.71\%) constitute the largest segments of the dataset. The final composition reflects both our discipline-balanced sampling of source papers and cross-disciplinary variation in the number of valid reasoning instances retained through the synthesis pipeline. Medicine \& Health (11.08\%) and Life Science (9.39\%) also contribute substantial portions, while the remaining disciplines, ranging from Materials Science \& Engineering (8.57\%) and Earth \& Atmospheric Science (8.45\%) to Energy \& Power (0.69\%), collectively ensure broad thematic coverage across diverse scientific domains.

Beyond disciplinary diversity, all four reasoning perspectives are well represented within the corpus. Mechanistic reasoning (35.6\%) and quantitative derivation (30.2\%) constitute the majority, aligning with the empirical nature of scientific papers, whereas boundary calibration (20.4\%) and hypothesis falsification (13.9\%) provide critical diversity in higher-order analytical evaluation.

The retention statistics in Table~\ref{tab:pipeline_yield} indicate that most generated instances pass the subsequent quality-control stages. In particular, the low proportion of rejections attributable to self-containment issues suggests that the reasoning skeleton helps reduce contextual dependencies more consistently during question generation.

\subsection{Dataset Analysis}

To evaluate the overall effectiveness of \oursdata{}, we benchmark it against five established scientific reasoning datasets, SCP-116K, TextbookReasoning, OpenScienceReasoning-2, NaturalReasoning, and MegaScience.

\paragraph{Length.}
\oursdata{} features the longest questions among all datasets, averaging 377.61 tokens (Table~\ref{tab:data_quality_analysis}). This length reflects the contextual information, variables, and boundary conditions often required to formulate well-posed scientific reasoning problems. The responses average 3139.45 tokens, ranking second overall. Together, these statistics indicate that \oursdata{} provides detailed reasoning supervision without relying on the longest response traces among the compared datasets.

\paragraph{Diversity.}
We quantify data diversity using the Vendi Score~\citep{friedman2022vendi} and Centroid Distance~\citep{suwanda2020analysis}. Despite its compact size (234K) compared with baselines containing up to 1.6M samples (Table~\ref{tab:data_quality_analysis}), \oursdata{} achieves the highest scores on both metrics (384.57 and 0.6183, respectively). These results indicate that \oursdata{} maintains high semantic diversity relative to the compared datasets and are consistent with the intended effect of our multi-perspective generation design.

\begin{table}[t]
  \centering\small\setlength{\tabcolsep}{4pt}
  \caption{Comparison of dataset length (average token length) and diversity.}
  \label{tab:data_quality_analysis}
  \resizebox{\columnwidth}{!}{%
  \begin{tabular}{l r cc cc}
    \toprule
    \multirow{2}{*}{\textbf{Dataset}} & \multirow{2}{*}{\textbf{Size}}
    & \multicolumn{2}{c}{\textbf{Length}}
    & \multicolumn{2}{c}{\textbf{Diversity}} \\
    \cmidrule(lr){3-4}\cmidrule(lr){5-6}
    & & \textbf{Q. Len.} & \textbf{Resp. Len.}
    & \textbf{Vendi} $\uparrow$ & \textbf{Cent. Dist.} $\uparrow$ \\
    \midrule
    SCP-116K & 274K & 122.09 & 555.68 & 173.42 & 0.4923 \\
    TextbookReasoning & 651K & 63.80 & 409.66 & 283.75 & 0.5598 \\
    NaturalReasoning & 1.1M & 76.28 & 766.33 & 349.42 & 0.6163 \\
    MegaScience & 1.2M & 170.77 & 692.93 & 373.78 & 0.6150 \\
    OpenScienceReasoning-2 & 1.6M & 217.40 & \textbf{5691.70} & 290.17 & 0.5610 \\
    \midrule
    \oursdata{}& 234K &\textbf{377.61} & 3139.45 & \textbf{384.57}& \textbf{0.6183} \\
    \bottomrule
  \end{tabular}
  }
\end{table}

\paragraph{Answer correctness.}
Given the dataset's cross-disciplinary breadth, comprehensive manual verification is prohibitively expensive. We therefore conduct large-scale automated verification on a random sample of 20K questions. For each question, Gemini-3.1 Pro\footnote{\url{https://ai.google.dev/gemini-api/docs/models/gemini-3.1-pro-preview}} independently generates a solution without access to the reference answer. GPT-OSS-120B~\citep{openai2025gptoss120bgptoss20bmodel} then judges whether the generated solution is semantically equivalent to the original response. This evaluation yields a 93.28\% consistency rate, providing additional evidence for answer reliability.

\begin{figure*}[t]
  \centering
  \includegraphics[width=0.85\linewidth]{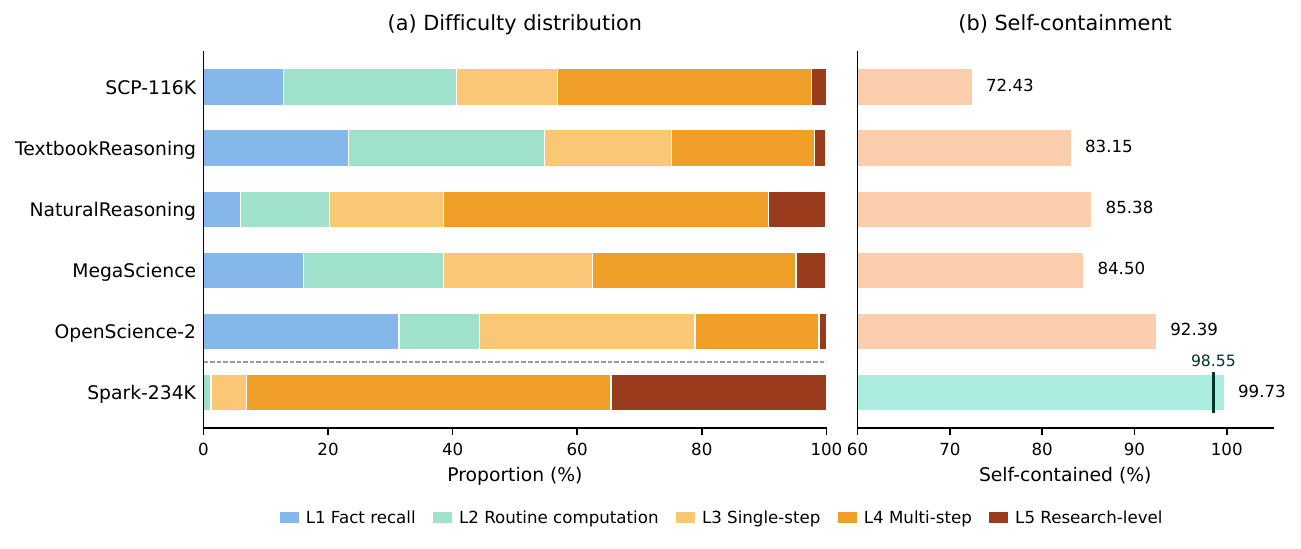}
  \caption{Difficulty distribution over five levels across datasets. \oursdata{} shifts mass toward higher difficulty, while baselines concentrate on low-difficulty recall or routine computation.}
  \label{fig:quality}
\end{figure*}

\paragraph{Difficulty.}
Figure~\ref{fig:quality}a highlights the limitations of existing datasets in eliciting deep reasoning. Using DeepSeek-V4-Flash, we grade 20K sampled questions per dataset on a five-level taxonomy (Appendix~\ref{sec:appendix:difficulty}). While baseline datasets contain high proportions of L1 (Fact recall) and L2 (Routine computation) questions, with their combined share exceeding 50\% in TextbookReasoning, \oursdata{} contains only 0.02\% L1 and 1.2\% L2 questions. Instead, over 93\% of our dataset requires high-order reasoning, dominated by L4 (Multi-step, 58.5\%) and L5 (Research-level, 34.6\%) problems. Notably, this distribution emerges without explicit difficulty-based filtering, suggesting that skeleton-guided generation naturally tends to produce reasoning-intensive questions.

\paragraph{Self-containment.}
Naively converting scientific papers into QA pairs often yields unusable questions that rely on unstated, external context. As Figure~\ref{fig:quality}b illustrates, multiple baseline datasets suffer from severe dependency issues, with their proportion of non-self-contained data exceeding 15\%. In supervised fine-tuning, introducing such a high ratio of contextually broken data is enough to severely degrade training performance, as it inadvertently teaches models to hallucinate missing variables rather than perform rigorous deduction. To isolate the architectural contribution of our reasoning skeleton in solving this structural flaw, we report \oursdata{} both before and after the final quality check. Even prior to the filtering stage, \oursdata{} achieves a 98.55\% self-containment rate, compared with 92.39\% for the strongest baseline. This result suggests that skeleton-guided generation substantially reduces context dependency at the generation stage. The final quality check further increases the self-containment rate to 99.73\%.

\paragraph{Human validation.}
To further verify the reliability of our automated quality assessments, we conduct a human validation study on 300 stratified samples across ten disciplines. Two paid domain experts for each discipline independently evaluate the samples across three dimensions: correctness, self-containment, and difficulty (Appendix~\ref{sec:appendix:difficulty}). Results show 90.3\% agreement with expert-derived answers, 97.7\% agreement on self-containment, and 94.3\% agreement on difficulty labels, confirming both the reliability of our annotations and the challenging nature of the dataset. Full details are provided in Appendix~\ref{sec:appendix:human_validation}.

\section{Experiments}
\label{sec:experiments}
%==============================================================================

\subsection{Experimental Setup}
\label{sec:setup}

\paragraph{Training.}
Fine-tuning is conducted on Llama3.1-8B-Base~\citep{grattafiori2024llama3herdmodels}, Qwen3-8B-Base, and Qwen3-14B-Base~\citep{yang2025qwen3technicalreport}. All comparisons share the same training hyperparameters; details are provided in Appendix~\ref{sec:appendix:training}.

\paragraph{Evaluation.}
For fair comparison, all models are evaluated in the zero-shot setting across two groups of benchmarks: \textbf{(1) general science reasoning benchmarks} (GPQA-Main [GPQA-M]~\citep{rein2024gpqa}, GPQA-Diamond [GPQA-D]~\citep{rein2024gpqa}, SuperGPQA [S-GPQA]~\citep{du2026supergpqa}, SciBench~\citep{wang2024scibench}, MMLU~\citep{hendrycks2020measuring}, and MMLU-Pro~\citep{wang2024mmlu}) and \textbf{(2) domain-specific benchmarks} (ChemBench~\citep{mirza2024large}, CS-Bench~\citep{song2025cs}, PubMedQA~\citep{jin2019pubmedqa}, MedQA-US~\citep{yao2024medqa}, GSM8K~\citep{cobbe2021training}, and MATH-500~\citep{hendrycksmath2021}). We report avg@3 or avg@5 for all benchmarks, with details provided in Appendix~\ref{sec:appendix:evaluation}.

\subsection{Main Results}
\label{sec:main_results}

Table~\ref{tab:main_results} reports the evaluation on six scientific reasoning benchmarks. The results empirically validate our central claim: preserving a paper's reasoning skeleton provides far denser supervision than the brute-force scaling of synthetic examples.

\begin{table*}[t]
  \centering\small\setlength{\tabcolsep}{5pt}
  \caption{Main results on six scientific reasoning benchmarks. Avg.\ is the macro average over all six. Within each model group, the best result per column is in \textbf{bold} and the second best is \underline{underlined}.}
  \label{tab:main_results}
  \begin{tabular}{l r @{\hspace{6pt}} ccc @{\hspace{6pt}} c @{\hspace{6pt}} cc @{\hspace{6pt}} c}
    \toprule
    & & \multicolumn{3}{c}{\textbf{Expert Science MCQ}} &
      \textbf{Quant.} &
      \multicolumn{2}{c}{\textbf{Broad Science}} & \\
    \cmidrule(lr){3-5}\cmidrule(lr){6-6}\cmidrule(lr){7-8}
    \textbf{Model / Training Data} & \textbf{Size}
      & \textbf{GPQA-M}& \textbf{GPQA-D} & \textbf{S-GPQA}
      & \textbf{SciBench}
      & \textbf{MMLU} & \textbf{MMLU-Pro}
      & \textbf{Avg.} \\
    \midrule
    \multicolumn{9}{l}{\textit{Llama3.1-8B-Base}} \\
    \quad Base                   & ---  & 24.55 & 24.75 & 14.41 & 17.41 & 54.06 & 24.39 & 26.60 \\
    \quad + SCP-116K             & 274K & 26.83 & 30.71 & 15.09 & 20.37 & 50.71 & 25.81 & 28.25 \\
    \quad + TextbookReasoning    & 651K & 30.11 & 33.95 & 24.76 & 31.83 & 66.25 & 47.73 & 39.11 \\
    \quad + NaturalReasoning     & 1.1M & 31.85 & 32.91 & 23.80 & 31.29 & 60.23 & 44.31 & 37.40 \\
    \quad + MegaScience          & 1.2M & \underline{33.71} & \textbf{38.89} & \underline{25.56} & \underline{36.87} & 68.79 & 49.79 & \underline{42.27} \\
    \quad + OpenScienceReasoning-2 & 1.6M & 29.91 & 26.77 & 24.09 & 27.59 & \textbf{70.10} & \underline{50.61} & 38.18 \\
    \rowcolor[HTML]{D6F7F0}
    \quad + \oursdata{} (ours)   & 234K  & \textbf{38.38} & \underline{37.37} & \textbf{28.47} & \textbf{43.80} & \underline{69.36} & \textbf{54.83} & \textbf{45.37} \\
    \midrule
    \multicolumn{9}{l}{\textit{Qwen3-8B-Base}} \\
    \quad Base                   & ---  & 32.59 & 29.29 & 29.05 & 30.30 & 66.22 & 45.24 & 38.78 \\
    \quad + SCP-116K             & 274K & 36.93 & 34.62 & 33.80 & 54.13 & 71.24 & 60.63 & 48.56 \\
    \quad + TextbookReasoning    & 651K & 40.75 & 42.90 & 37.01 & 61.31 & 73.07 & 64.26 & 53.22 \\
    \quad + NaturalReasoning     & 1.1M & 40.03 & 41.29 & 34.44 & 59.83 & 72.15 & 61.91 & 51.61 \\
    \quad + MegaScience          & 1.2M & \underline{43.75} & \underline{44.95} & \underline{38.66} & \underline{64.83} & 79.65 & 66.31 & \underline{56.36} \\
    \quad + OpenScienceReasoning-2 & 1.6M & 40.78 & 39.36 & 38.55 & 62.37 & \underline{81.15} & \underline{69.55} & 55.29 \\
    \rowcolor[HTML]{D6F7F0}
    \quad + \oursdata{} (ours)   & 234K  & \textbf{50.22} & \textbf{52.53} & \textbf{41.84} & \textbf{66.35} & \textbf{81.50} & \textbf{72.54} & \textbf{60.83} \\
    \midrule
    \multicolumn{9}{l}{\textit{Qwen3-14B-Base}} \\
    \quad Base                   & ---  & 37.95 & 38.38 & 33.50 & 50.16 & 76.01 & 59.35 & 49.23 \\
    \quad + SCP-116K             & 274K & 42.80 & 43.75 & 35.29 & 59.29 & 77.37 & 64.99 & 53.92 \\
    \quad + TextbookReasoning    & 651K & 44.27 & 46.81 & 42.15 & 63.61 & 79.25 & 68.53 & 57.44 \\
    \quad + NaturalReasoning     & 1.1M & 44.09 & 44.83 & 34.16 & 63.27 & 77.59 & 67.94 & 55.31 \\
    \quad + MegaScience          & 1.2M & \underline{47.32} & \underline{48.99} & \underline{44.47} & \underline{67.81} & 82.44 & 71.65 & \underline{60.45} \\
    \quad + OpenScienceReasoning-2 & 1.6M & 46.64 & 47.83 & 44.21 & 65.41 & \textbf{83.99} & \underline{72.20} & 60.05 \\
    \rowcolor[HTML]{D6F7F0}
    \quad + \oursdata{} (ours)   & 234K  & \textbf{53.71} & \textbf{55.05} & \textbf{47.62} & \textbf{69.55} & \underline{83.65} & \textbf{73.68} & \textbf{63.88} \\
    \bottomrule
  \end{tabular}
\end{table*}

\paragraph{Data Efficiency.}
The macro averages highlight the data efficiency of our approach. Across all three backbones, \oursdata{} consistently achieves the highest overall performance using only 234K examples---less than one-fifth the volume of million-scale baselines such as MegaScience (1.2M). For instance, fine-tuning Qwen3-8B on \oursdata{} yields an average score of 60.83, compared with 56.36 for the strongest baseline. These results show that \oursdata{} achieves strong downstream performance with substantially fewer training examples, highlighting the importance of data construction quality in addition to dataset scale.

\paragraph{Performance on Challenging Reasoning Benchmarks.}
The performance gains are particularly pronounced on benchmarks requiring complex, multi-step reasoning. On the Expert Science MCQ cluster (GPQA series) and the quantitative SciBench, \oursdata{} consistently achieves strong performance relative to the compared datasets. For Qwen3-14B, training on \oursdata{} yields GPQA-M and GPQA-D scores of 53.71 and 55.05, respectively, outperforming models trained on substantially larger corpora. Improvements on MMLU-Pro show a similar trend. These results are consistent with the high proportion of L4 and L5 questions observed in \oursdata{}, suggesting that skeleton-guided synthesis provides effective supervision for challenging reasoning tasks.

\subsection{Domain-Specialized Results}
\label{sec:domain_results}

Table~\ref{tab:domain_results} evaluates performance across distinct scientific disciplines. \oursdata{} consistently achieves the highest macro averages across all three backbones, indicating broad improvements across the evaluated chemistry, computer science, medicine, and mathematics benchmarks.

Notably, \oursdata{} maintains a substantial advantage on domain-specific benchmarks such as MedQA-US (e.g., 79.36 vs.\ 72.58 for MegaScience on Qwen3-14B). This result suggests that the gains from our training data transfer effectively to specialized scientific domains.

\begin{table*}[t]
  \centering\small\setlength{\tabcolsep}{4pt}
  \caption{Domain-specialized transfer results across chemistry, computer science, medicine, and mathematics.}
  \label{tab:domain_results}
  \begin{tabular}{l r c c c c c c c}
    \toprule
    \textbf{Model / Training Data} & \textbf{Size}
      & \textbf{ChemBench} & \textbf{CS-Bench}
      & \textbf{PubMedQA} & \textbf{MedQA-US}
      & \textbf{GSM8K} & \textbf{MATH-500}
      & \textbf{Avg.} \\
    \midrule
    \multicolumn{9}{l}{\textit{Llama3.1-8B-Base}} \\
    \quad Base                   & ---  & 23.03 & 49.90 & 51.20 & 48.78 & 56.40 & 19.20 & 41.42 \\
    \quad + MegaScience          & 1.2M & \underline{42.97} & \underline{59.89} & \textbf{76.60} & \underline{62.03} & \underline{71.65} & \underline{45.20} & \underline{59.72} \\
    \rowcolor[HTML]{D6F7F0}
    \quad + \oursdata{} (ours)   & 234K  & \textbf{44.51} & \textbf{64.29} & \underline{74.20} & \textbf{66.35} & \textbf{85.10} & \textbf{64.00} & \textbf{66.41} \\
    \midrule
    \multicolumn{9}{l}{\textit{Qwen3-8B-Base}} \\
    \quad Base                   & ---  & 49.60 & 69.16 & 61.20 & 55.54 & 88.63 & 70.80 & 65.82 \\
    \quad + MegaScience          & 1.2M & \underline{53.01} & \underline{76.07} & \textbf{77.40} & \underline{66.93} & \underline{93.03} & \underline{79.60} & \underline{74.34} \\
    \rowcolor[HTML]{D6F7F0}
    \quad + \oursdata{} (ours)   & 234K  & \textbf{55.85} & \textbf{77.61} & \underline{75.60} & \textbf{75.33} & \textbf{94.09} & \textbf{81.20} & \textbf{76.61} \\
    \midrule
    \multicolumn{9}{l}{\textit{Qwen3-14B-Base}} \\
    \quad Base                   & ---  & 53.73 & 75.98 & 77.00 & 63.71 & 93.86 & 71.60 & 72.65 \\
    \quad + MegaScience          & 1.2M & \textbf{58.82} & \underline{79.92} & \textbf{78.80} & \underline{72.58} & \underline{94.77} & \underline{82.00} & \underline{77.82} \\
    \rowcolor[HTML]{D6F7F0}
    \quad + \oursdata{} (ours)   & 234K  & \underline{57.63} & \textbf{81.50} & \underline{78.20} & \textbf{79.36} & \textbf{94.82} & \textbf{85.37} & \textbf{79.48} \\
    \bottomrule
  \end{tabular}
  \vspace{-1mm}
\end{table*}

\subsection{Data Efficiency and Scaling}
\label{sec:data_scaling}
\begin{figure}[t]
  \centering
  \includegraphics[width=0.9\linewidth]{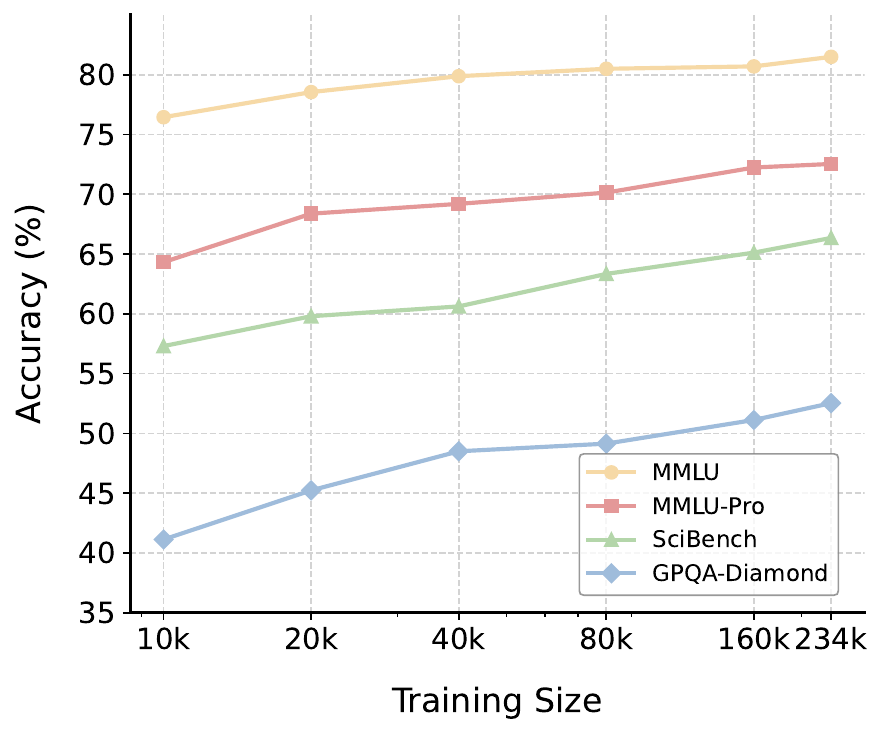}
  \caption{Performance scaling with the number of synthesized training records from \oursdata{}.}
  \label{fig:data_scaling}
\end{figure}

All experiments in this and the following ablation use Qwen3-8B-Base, evaluated on MMLU, MMLU-Pro, SciBench, and GPQA-Diamond under the same training and evaluation protocol as our main experiments. Here we train on progressively larger random subsets of \oursdata{} to measure how performance scales with the amount of synthesized data (Figure~\ref{fig:data_scaling}).

While the total dataset volume is relatively compact, the scaling curve exhibits a consistent and incremental improvement across benchmarks as the data size expands. This steady upward trajectory indicates that the model has not yet saturated on the provided scientific reasoning patterns. It suggests a highly promising direction for future work, where simply applying our automated extraction pipeline to a broader corpus of academic literature could naturally yield further scaling gains.

\subsection{Pipeline Ablation}
\label{sec:ablation}
We ablate the two central designs of \ours{}, with all settings trained on a fixed 40K set (Table~\ref{tab:ablation}).

\paragraph{Unit of synthesis.}
We replace skeleton extraction with three alternatives that feed the generator different representations of the same paper: the whole paper, chunked passages, or an LLM-generated summary. The rest of the pipeline remains unchanged. All baseline representations degrade sharply. Whole-paper input performs worst ($-9.15$), as processing a long document diffuses the central claims and yields shallow questions. Chunking and summarization also lose ground ($-5.63$, $-6.71$) by severing cross-section claim--evidence chains or discarding the reasoning skeleton entirely. In contrast, extracting a reasoning skeleton preserves the paper's claim--evidence--derivation trajectory, yielding the highest-quality synthetic data.

\paragraph{Question generation.}
We fix the skeleton and perform leave-one-out ablations over the four perspectives, resampling from the remaining three to maintain 40K examples so that variants differ only in the included perspectives. With data volume fixed, the performance gaps are smaller, yet removing any perspective still causes a clear drop, indicating that each contributes complementary signal. Quantitative derivation has the largest effect ($-4.64$), likely because it contributes many computation-intensive questions that strengthen the model's quantitative reasoning ability.

\begin{table}[t]
  \centering\small\setlength{\tabcolsep}{4pt}
  \caption{Ablation of key \ours{} design choices.}
  \label{tab:ablation}
  \begin{tabular}{lcc}
    \toprule
    \textbf{Setting} & \textbf{Avg.} & \textbf{$\Delta$} \\
    \midrule
    \ours{} (full)        & 63.46 & --- \\
    \midrule
    \multicolumn{3}{l}{\textit{Unit of synthesis}} \\
    \quad w/o Skeleton (\textrightarrow{} whole paper) & 54.31 & $-9.15$ \\
    \quad w/o Skeleton (\textrightarrow{} chunks)      & 57.83 & $-5.63$ \\
    \quad w/o Skeleton (\textrightarrow{} summary)     & 56.75 & $-6.71$ \\
    \midrule
    \multicolumn{3}{l}{\textit{Question generation}} \\
    \quad w/o Mechanistic    & 61.59 & $-1.87$ \\
    \quad w/o Falsification  & 60.30 & $-3.16$ \\
    \quad w/o Quantitative   & 58.82 & $-4.64$ \\
    \quad w/o Boundary       & 61.19 & $-2.27$ \\
    \bottomrule
  \end{tabular}
\end{table}

%==============================================================================
\section{Related Work}
\label{sec:related}
%==============================================================================
\subsection{Scientific Reasoning Data}
\label{sec:related:reasoning_data}
Scientific post-training data has expanded from curricular sources to large synthetic mixtures. TextbookReasoning~\citep{fan2025megascience} extracts QA from university textbooks, while domain-specific efforts such as ChemData700K~\citep{zhang2024chemllm} tune models on field-focused instruction data. Broader collections, including SCP-116K~\citep{lu2025scp}, OpenScienceReasoning-2~\citep{hf_opensciencereasoning2}, NaturalReasoning~\citep{yuan2026naturalreasoning}, and MegaScience~\citep{fan2025megascience}, improve scale and coverage. However, scale alone does not guarantee depth. Textbooks favor settled knowledge, while synthetic mixtures often default to factual recall and formulaic problem-solving. This motivates sourcing supervision from texts rich in scientific reasoning, rather than mere facts.

\subsection{Data Synthesis from Scientific Literature}
\label{sec:related:literature_synthesis}
Frontier papers naturally fulfill this requirement, yet scalable task conversion remains nontrivial. While human annotations~\citep{dasigi2021dataset} resist scaling, LLM pipelines now synthesize tasks from full papers~\citep{wan2024sciqag,liu2026wildsci}, chunked text~\citep{sarthi2024raptorrecursiveabstractiveprocessing}, or retrieved passages~\citep{wang2025researchgpt}. However, treating a paper as a monolithic document diffuses its core claims, while arbitrary chunking severs vital cross-section evidence chains. \ours{} instead distills a paper's claims, evidence, and derivations into a compact \emph{reasoning skeleton} as source, yielding self-contained, mechanism-oriented questions.

\section{Conclusion}
We introduced \ours{} to address the limitations of existing scientific reasoning data by treating a paper's \emph{claim--evidence--derivation} structure as the fundamental unit of synthesis. \ours{} distills papers into reasoning skeletons, generates questions across four scientific perspectives, and enforces strict consistency. The resulting \oursdata{} offers greater difficulty and diversity, and experiments demonstrate it substantially improves the scientific reasoning capabilities of base models.
%==============================================================================
\section*{Limitations}
%==============================================================================
This work has two main limitations. First, constrained by computational resources, our experiments are limited to supervised fine-tuning; we leave the use of \oursdata{} for reinforcement learning, including more verifiable reward design, to future work. Second, given the scale and disciplinary breadth of \oursdata{}, comprehensive expert verification of the entire dataset is prohibitively expensive. We therefore rely primarily on automated consistency checks for large-scale quality control, complemented by a stratified human validation study on 300 samples across all ten disciplines. Extending expert evaluation to a substantially larger portion of the dataset would provide a more comprehensive assessment of data reliability.

\section*{Acknowledgements}
This work was supported by Shanghai Artificial Intelligence Laboratory.

%==============================================================================
\bibliography{custom}
%==============================================================================
\clearpage
\newpage

\appendix

%==============================================================================
\section{The Source Corpus: \corpus{}}
%==============================================================================
\label{sec:appendix:scibase}

\corpus{} serves as the foundational corpus of open-access scientific literature for our dataset, comprising over 3.36 million parsed and formula-preserving papers with coverage up to March 2026. To ensure strict data quality for downstream synthesis, we applied rigorous filtering criteria to the raw aggregated documents, explicitly removing incomplete papers, such as those missing abstracts or essential metadata.

As detailed in Table~\ref{tab:corpus_stats}, the finalized high-quality corpus spans ten major scientific disciplines. While the distribution inherently skews towards Medicine \& Health Sciences and Life Sciences (collectively accounting for 61.3\%) due to natural open-access publishing volumes, this massive, structured repository provides a highly reliable and diverse pool for our discipline-aware sampling pipeline.

\begin{table}[h]
  \centering\small
  \caption{Disciplinary distribution of \corpus{}.}
  \label{tab:corpus_stats}
  \begin{tabular}{lrr}
    \toprule
    \textbf{Discipline} & \textbf{\# Papers} & \textbf{\%} \\
    \midrule
Medicine and Health Sciences     & 1{,}303{,}682 & 38.8\% \\
Life Sciences                    & 756{,}011    & 22.5\% \\
Engineering and Mfg.\ Science    & 399{,}840    & 11.9\% \\
Earth and Atmospheric Sciences   & 201{,}607    & 6.0\% \\
Physics                          & 181{,}431    & 5.4\% \\
Chemistry                        & 181{,}446    & 5.4\% \\
Math and Computational Science   & 134{,}452    & 4.0\% \\
Materials Science and Eng.       & 120{,}963    & 3.6\% \\
Astronomy and Space Sciences     & 63{,}848     & 1.9\% \\
Energy and Power Science         & 16{,}800     & 0.5\% \\
    \midrule
    \textbf{Total}                   & 3.36M & 100\% \\
    \bottomrule
  \end{tabular}
\end{table}

% ===== 放到附录 =====
\section{Construction Funnel}
\label{sec:appendix:funnel}

Table~\ref{tab:pipeline_yield} tracks the volume dynamics and rejection rates across the different stages of our data synthesis pipeline.

\paragraph{Skeleton Extraction.}
The pipeline starts from approximately 370K papers obtained after corpus-level filtering. Of these, roughly 325K are retained after skeleton extraction and lightweight post-processing. The approximately 45K excluded papers fall into three main categories. First, for some papers, no clear target conclusion could be identified, including documents without an explicit conclusion (e.g., certain survey papers) as well as noisy or misclassified entries in the original \corpus{} corpus, such as speeches or lecture transcripts. Second, some papers failed to yield a valid skeleton after three API retries, typically due to transient network issues or unusually long or short source documents. Third, we apply several lightweight heuristic checks to the extracted skeletons: we discard skeletons containing three or fewer argumentation steps, as well as those that explicitly depend on figures or tables, in order to preserve sufficient reasoning structure and self-containedness.

\paragraph{QA Filtering and Finalization.}
The skeleton-grounded check removes 14.97\% of the generated QA pairs. Among the QA pairs rejected at this stage, 7.92\% are rejected because the questions are not self-contained, suggesting that the upstream skeleton extraction and post-processing stages effectively reduce contextual dependency issues. In contrast, 65.87\% of the rejected QA pairs are filtered because their answers are inconsistent with the original paper skeleton. This distribution indicates that the quality auditor primarily removes answers that deviate from the source reasoning, rather than questions with insufficient contextual information.

In the subsequent deduplication and decontamination stage, an additional 3.04\% of the remaining QA pairs are removed, yielding approximately 234K final instances. These removals arise from semantic duplication, while no instance triggers our 13-gram exact-match decontamination criterion against the downstream evaluation benchmarks considered in this work. This observation indicates low exact lexical overlap with these evaluation sets, reducing the risk of direct benchmark leakage under our decontamination criterion.

\begin{table}[t]
\centering
\caption{
Construction funnel of \oursdata{}.
The retained size denotes the number of instances remaining after each stage.
$^\dagger$Percentages for the four rejection reasons denote their respective
shares among the QA pairs rejected during the skeleton-grounded check,
rather than removal rates over the full dataset.
}
\small
\resizebox{\columnwidth}{!}{%
\begin{tabular}{lcc}
\toprule
\textbf{Stage} & \textbf{Retained Size} & \textbf{Removal Rate} \\
\midrule
Initial sampling              & 370K & -- \\
Skeleton extraction           & 325K & -- \\
Question generation           & 293K & -- \\
Answer generation             & 283K & -- \\
Skeleton-grounded check       & 241K & 14.97\% \\
\quad Question not self-contained
                              & -- & 7.92\%$^\dagger$ \\
\quad Question not paper-grounded
                              & -- & 13.61\%$^\dagger$ \\
\quad Answer inconsistent with skeleton
                              & -- & 65.87\%$^\dagger$ \\
\quad Answer internally incoherent
                              & -- & 12.60\%$^\dagger$ \\
Deduplication \& decontamination
                              & 234K & 3.04\% \\
\bottomrule
\end{tabular}}
\label{tab:pipeline_yield}
\end{table}

\begin{figure}[t]
  \centering
  \includegraphics[width=\linewidth]{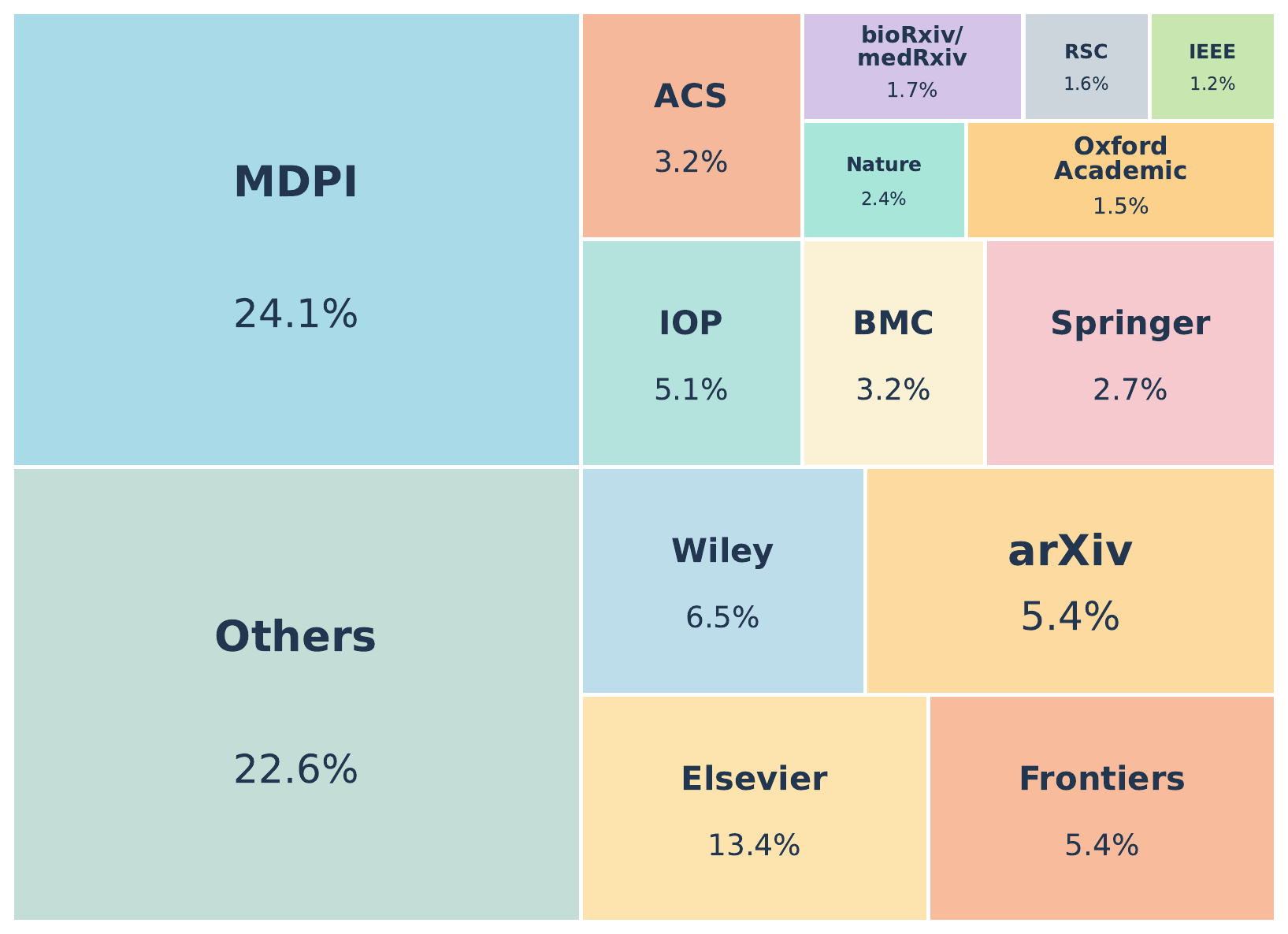}
  \caption{Publisher distribution of the source literature underlying \oursdata{}.}
  \label{fig:publisher_treemap}
\end{figure}

\paragraph{Source Provenance and Data Diversity.}
To characterize the composition of the finalized dataset, Figure~\ref{fig:publisher_treemap} illustrates the distribution of source publishing venues and platforms within \oursdata{}. The underlying corpus aggregates literature from a heterogeneous mix of prominent academic publishers and preprint repositories. As shown in the treemap, the dataset draws substantially from major venues including MDPI (24.1\%), Elsevier (13.4\%), Wiley (6.5\%), arXiv (5.4\%), and Frontiers (5.4\%). This is supplemented by professional societies, publishers, and specialized platforms, including IOP (5.1\%), ACS (3.2\%), BMC (3.2\%), Springer (2.7\%), Nature Portfolio (2.4\%), bioRxiv/medRxiv (1.7\%), RSC (1.6\%), Oxford Academic (1.5\%), and IEEE (1.2\%), alongside a long-tail category of other venues comprising 22.6\%.

This broad source provenance provides a diverse foundation for the construction of \oursdata{}. The included publishers, academic societies, and preprint repositories span heterogeneous disciplinary communities, vocabulary conventions, and scientific writing practices. Drawing source reasoning structures from this diverse collection reduces dependence on the conventions of any single venue or publisher and broadens the linguistic and disciplinary coverage of the synthesized reasoning instances.

%==============================================================================
\section{Training Details}
\label{sec:appendix:training}
%==============================================================================

We conduct fine-tuning using LlamaFactory~\citep{zheng2024llamafactory}. To ensure rigorous evaluation integrity, all baseline datasets undergo a strict 13-gram exact match decontamination process against all downstream test benchmarks prior to training. Unless otherwise specified, experiments utilize 16 GPUs with a per-device batch size of 4 and 2 gradient accumulation steps. For 32-GPU runs, we adjust gradient accumulation to 1 to maintain a consistent global batch size. Table~\ref{tab:training_details} summarizes the remaining shared hyperparameters.

\begin{table}[h]
  \centering\small
  \caption{Training Configuration.}
  \label{tab:training_details}
  \begin{tabular}{lc}
    \toprule
    \textbf{Configuration} & \textbf{Value} \\
    \midrule
    Cutoff length & 32{,}768 \\
    Epochs & 3.0 \\
    Learning rate & 5.0e-6 \\
    Warmup ratio & 0.05 \\
    Per-device batch size & 4 \\
    Gradient accumulation (16 GPUs) & 2 \\
    Gradient accumulation (32 GPUs) & 1 \\
    Flash attention & FlashAttention-2 \\
    Liger Kernel & Enabled \\
    Thinking mode & Enabled \\
    Average tokens across devices & True \\
    Drop last dataloader batch & False \\
    \bottomrule
  \end{tabular}
\end{table}

%==============================================================================
\section{Evaluation Details}
\label{sec:appendix:evaluation}
%==============================================================================

We organize evaluation benchmarks into two groups. The first group contains general science reasoning benchmarks used in the main comparison, covering expert science QA, quantitative scientific problem solving, and broad multidisciplinary science understanding. The second group contains domain-specific benchmarks for chemistry, computer science, medicine, and mathematics, which evaluate whether scientific fine-tuning transfers to specialized fields. Table~\ref{tab:evaluation_details} summarizes all evaluation sets.

For evaluation implementation, we use the same codebase as MegaScience\footnote{\url{https://github.com/GAIR-NLP/lm-open-science-evaluation}}, which standardizes prompting, decoding, and answer extraction across benchmarks. All models are evaluated in the zero-shot setting on 8 GPUs with temperature set to 0, and reported results are avg@3 or avg@5 depending on the benchmark.

\begin{table*}[t]
  \centering\small\setlength{\tabcolsep}{5pt}
  \caption{Benchmark details for our evaluation.}
  \label{tab:evaluation_details}
  \begin{tabular}{>{\centering\arraybackslash}m{2.6cm}
                  >{\centering\arraybackslash}m{1.4cm}
                  >{\centering\arraybackslash}m{1.3cm}
                  >{\centering\arraybackslash}m{3.0cm}
                  >{\arraybackslash}p{5.1cm}}
    \toprule
    \textbf{Benchmark} & \textbf{Size} & \textbf{Report} & \textbf{Category} & \textbf{Description} \\
    \midrule
    \multicolumn{5}{c}{\textit{General science reasoning benchmarks}} \\
    \midrule
    GPQA-Main\\~\citep{rein2024gpqa} & 448 & avg@5 & Expert science QA & Graduate-level, Google-proof multiple-choice questions across biology, chemistry, and physics. \\
    GPQA-Diamond\\~\citep{rein2024gpqa} & 198 & avg@5 & Expert science QA & The higher-confidence GPQA subset selected for expert agreement and difficulty. \\
    SuperGPQA\\~\citep{du2026supergpqa} & 26{,}529 & avg@3 & Broad graduate science & Large-scale graduate-level benchmark spanning many disciplines and subfields. \\
    SciBench\\~\citep{wang2024scibench} & 640 & avg@5 & Quantitative science & Aggregated evaluation over the tested chemistry, mathematics, and physics subsets. \\
    MMLU\\~\citep{hendrycks2020measuring} & 14{,}042 & avg@3 & Broad knowledge & Multitask benchmark covering broad academic and professional subjects, including science. \\
    MMLU-Pro\\~\citep{wang2024mmlu} & 12{,}032 & avg@3 & Broad knowledge & More challenging multitask benchmark with harder options and stronger reasoning demands. \\
    \midrule
    \multicolumn{5}{c}{\textit{Domain-specific benchmarks}} \\
    \midrule
    ChemBench\\~\citep{mirza2024large} & 2{,}788 & avg@5 & Chemistry & Aggregated evaluation over the tested multiple-choice and string-match subsets. \\
    CS-Bench\\~\citep{song2025cs} & 1{,}778 & avg@5 & Computer science & Aggregated evaluation over the tested assertion and multiple-choice subsets. \\
    PubMedQA\\~\citep{jin2019pubmedqa} & 500 & avg@5 & Medicine & Biomedical question answering over PubMed-derived contexts. \\
    MedQA-US\\~\citep{yao2024medqa} & 1{,}273 & avg@5 & Medicine & Medical licensing-style multiple-choice questions from the US subset. \\
    GSM8K\\~\citep{cobbe2021training} & 1{,}319 & avg@5 & Mathematics & Grade-school math word problems requiring multi-step arithmetic reasoning. \\
    MATH-500\\~\citep{hendrycksmath2021} & 500 & avg@5 & Mathematics & A 500-problem subset of competition-style mathematical reasoning problems. \\
    \bottomrule
  \end{tabular}
\end{table*}

\section{Baseline Data Selection}
\label{sec:appendix:baseline_data_selection}

\subsection{Baseline Data Selection Criteria}

For baseline comparison, we carefully select the most popular scientific SFT datasets currently available in the open-source community. The selection criteria prioritize datasets that are (1) widely adopted in prior work, (2) publicly accessible, and (3) contain complete CoT reasoning traces suitable for SFT. Based on these criteria, we include SCP-116K~\citep{lu2025scp} (we adopt the latest version, which contains 274K samples; the original authors did not rename the dataset despite the increased size), TextbookReasoning~\citep{fan2025megascience}, NaturalReasoning~\citep{yuan2026naturalreasoning}, MegaScience~\citep{fan2025megascience}, and OpenScienceReasoning-2~\citep{hf_opensciencereasoning2}. For all these datasets, we directly use their full original data with the original CoT in our experiments.

\subsection{Controlled Comparison with WildSci}

WildSci~\citep{liu2026wildsci} is designed for reinforcement learning from verifiable rewards (RLVR). Although it provides verified multiple-choice answers, it lacks the reasoning traces required for SFT, which precludes a direct comparison in Table~\ref{tab:main_results}. To enable a more detailed and fair evaluation, we augment WildSci with reasoning responses and present controlled comparisons below.

We first apply n-gram decontamination against all evaluation benchmarks. For each remaining question, Qwen3.5-27B, the answer model used by SPARK, attempts up to four times to generate a valid reasoning response. We retain a response only when its final answer matches WildSci's gold option and discard questions for which no valid response is obtained. We then sample 20K instances from the filtered pool.

We conduct SFT on Qwen3-8B-Base under identical settings, and present the results in Table~\ref{tab:controlled_sft_comparison}.

\begin{table}[H]     % 把 [htbp] 换成 [H]
\centering
\caption{Spark-234K vs. WildSci SFT comparison.}
\label{tab:controlled_sft_comparison}
\resizebox{\columnwidth}{!}{%
\begin{tabular}{lcccccc}
\hline
\textbf{Training Data} & \textbf{Size} & \textbf{GPQA-M} & \textbf{GPQA-D} & \textbf{S-GPQA} & \textbf{MMLU-Pro} & \textbf{Avg.} \\
\hline
WildSci (CoT) & 20K & 44.66 & 43.49 & 37.69 & 65.05 & 47.72 \\
Spark-234K (subset) & 20K & 46.63 & 45.37 & 39.75 & 68.31 & 50.02 \\
\hline
\end{tabular}%
}
\end{table}

As shown in Table~\ref{tab:controlled_sft_comparison}, Spark-234K outperforms the SFT-adapted WildSci data on all four overlapping benchmarks, with an average improvement of 2.30 points (50.02 vs. 47.72). Together with the intrinsic comparison presented in the main paper, these results support the benefit of SPARK's skeleton-guided pipeline for question synthesis and answer verification.

Beyond the performance comparison discussed above, our SPARK pipeline also differs from WildSci in several key design aspects:
\begin{itemize}
\item \textbf{Paper processing.} WildSci processes the full paper directly, whereas SPARK first extracts a paper-level reasoning skeleton and generates questions from the resulting evidence trajectory.
\item \textbf{Question generation.} WildSci generates three multiple-choice questions per paper. SPARK generates at most one question per perspective and skips perspectives unsupported by the source evidence, adapting its questions to each paper's content.
\item \textbf{Answer verification.} WildSci uses model voting to identify unanswerable or divergent items, whereas SPARK verifies generated answers against the extracted reasoning skeleton (Section~\ref{sec:method:quality}).
\end{itemize}

%==============================================================================
\section{Pipeline Details}
\label{sec:appendix:pipeline_details}
%==============================================================================

\subsection{Model and Decoding Configuration}
\label{sec:appendix:pipeline_models}

Table~\ref{tab:pipeline_api_config} summarizes the API configuration used for skeleton extraction and multi-perspective question generation. Both stages call DeepSeek-V4-Flash with thinking enabled and high reasoning effort.

\begin{table}[H]
  \centering\small
  \caption{API configuration for skeleton extraction and question generation.}
  \label{tab:pipeline_api_config}
  \begin{tabular}{lc}
    \toprule
    \textbf{Configuration} & \textbf{Value} \\
    \midrule
    Model & DeepSeek-V4-Flash \\
    Thinking & Enabled \\
    Reasoning effort & High \\
    \bottomrule
  \end{tabular}
\end{table}

Answer generation and skeleton-grounded quality checking are performed with locally deployed models. Qwen3.5-27B is used to generate answers, while GPT-OSS-120B is used as the quality checker. Table~\ref{tab:pipeline_local_config} reports the decoding configuration for both stages.

\begin{table}[H]
  \centering\small\setlength{\tabcolsep}{5pt}
  \caption{Local model configuration for answer generation and skeleton-grounded quality checking.}
  \label{tab:pipeline_local_config}
  \begin{tabular}{lcc}
    \toprule
    \textbf{Configuration} & \textbf{Answer generation} & \textbf{Quality check} \\
    \midrule
    Model & Qwen3.5-27B & GPT-OSS-120B \\
    Max new tokens & 32{,}768 & 16{,}384 \\
    Max input tokens & 2048 & 32{,}768 \\
    Temperature & 1.0 & 0.2 \\
    Top-$p$ & 0.95 & 0.95 \\
    Top-$k$ & 20 & 20 \\
    Min-$p$ & 0.0 & 0.0 \\
    Presence penalty & 1.5 & 1.5 \\
    Repetition penalty & 1.0 & 1.0 \\
    \bottomrule
  \end{tabular}
\end{table}

\subsection{Prompt Templates}
\label{sec:appendix:pipeline_prompts}

We provide the prompt templates used in the pipeline. Skeleton extraction uses two prompts: Figure~\ref{fig:prompt_conclusion_location} localizes the core conclusion from the title and abstract, and Figure~\ref{fig:prompt_trajectory_extraction} reconstructs the claim-conditioned trajectory from the full paper content.

Question generation is implemented as a modular prompt. Figure~\ref{fig:prompt_question_generation} gives the shared prompt framework, which defines the global objective, self-containment constraints, skip conditions, and output schema. The perspective-specific content is injected through the placeholder \texttt{\{perspective\_guidance\}}: at runtime, it is replaced by one of the four definitions in Figure~\ref{fig:prompt_perspective_guidance}. This design keeps the output format and global quality constraints fixed while allowing each generated question to target a different reasoning role, such as mechanism explanation, hypothesis falsification, quantitative derivation, or boundary calibration. Finally, Figure~\ref{fig:prompt_answer_generation} shows the answer-generation prompt, and Figure~\ref{fig:prompt_quality_check} shows the skeleton-grounded quality-check prompt used to filter QA pairs.

\begin{figure*}[t]
\begin{promptbox}{\textit{Prompt for Core-Conclusion Localization}}
\begin{lstlisting}[style=prompt]
### Role
You are a senior scientific reviewer identifying the central claim of a research paper.

### Task
Read the title and abstract and articulate the paper's core conclusion. Capture the main finding precisely; where a single sentence cannot do it justice, use the supporting fields to record the qualifications, scope, or significance that make the claim complete.

### Guidance
- The headline conclusion must be a concrete scientific assertion, not a topic description. Prefer "X increases Y under condition Z" over "this work studies the relationship between X and Y".
- Many findings are only meaningful with their conditions, mechanism, or scope attached. Do not amputate these to fit one sentence; place them in the supporting fields instead.
- Record only what the abstract actually commits to. Do not invent results, and do not promote previewed methods into findings.
- If there is no clear scientific argument (e.g. an editorial, news piece, abstract collection, or purely descriptive catalog without a central thesis), output an empty JSON object {{}}.

### Output Schema
Return ONLY a JSON object with this exact structure:
{{
  "core_conclusion": "The single most central finding, stated as one concrete sentence.",
  "key_qualifications": ["A condition, scope limit, or boundary under which the conclusion holds"],
  "significance": "Why this finding matters or what it resolves, in one sentence (empty string if the abstract does not state it)."
}}

### Critical Rules
- core_conclusion must be one substantive, self-contained sentence naming the specific subject and the asserted effect or mechanism. Keep it focused; offload conditions and scope to key_qualifications rather than overloading this sentence.
- key_qualifications captures the decisive conditions, parameters, or limits that the conclusion depends on. Use an empty list only when the finding is genuinely unconditional.
- significance is a single sentence and may be an empty string if not stated.
- If the conclusion is vague, hedged beyond recovery, or absent, output {{}}.
- Output ONLY the raw JSON object. Do NOT wrap it in markdown formatting or ```json code fences. No explanatory prose.
- Do not refer to "the paper", "the authors", "the study", sections, figures, or tables.

### Title
{paper_title}

### Abstract
{paper_abstract}
\end{lstlisting}
\end{promptbox}
\caption{Prompt for Core-Conclusion Localization.}
\label{fig:prompt_conclusion_location}
\end{figure*}

\begin{figure*}[t]
\begin{promptbox}{\textit{Prompt for Claim-Conditioned Trajectory Extraction}}
\begin{lstlisting}[style=prompt]
### Role
You are a senior scientific reviewer reconstructing the logical structure that supports a paper's stated conclusion.

### Task
The core conclusion has already been established. Treat it as the fixed endpoint of the argument. Your work has two parts:
1. Read the full content and reconstruct the step-by-step argumentation chain that leads to this conclusion, identifying key logical steps (not sections) and focusing on how the conclusion is reached through experiments, observations, derivations, or reasoning.
2. Judge which training perspectives the reconstructed trajectory genuinely supports, using the definitions below.

### Core Conclusion (fixed target)
{core_conclusion}

### Training Perspectives
Match perspectives to the nature of the trajectory itself:
- mechanistic_reasoning -> the work centers on a new mechanism or causal account of why/how something happens.
- hypothesis_falsification -> the work hinges on testing competing explanations and ruling some out.
- quantitative_derivation -> the work is computation-heavy, rich in values, units, and formulas.
- boundary_calibration -> the work is careful about what its evidence does and does not justify.

### Critical Rules
- steps must capture the logical flow from evidence to the given conclusion. Do NOT produce a section-by-section summary.
- Each step must describe one logical unit of the argument (an experiment that tests a hypothesis, a comparison that rules out an alternative, a derivation that connects observations to a mechanism, etc.).
- The "description" MUST explicitly mention the specific experimental conditions, chemical names, mathematical operators, or equipment setups used. Avoid generic phrasing like "an analysis was conducted".
- The "observables" MUST contain concrete anchors: exact numerical values with units, specific formulas/variables, statistical bounds, key entities, or distinct phenomenological signs. Never write placeholder text like "did an experiment and got a result".
- recommended_perspectives: list 1-4 perspectives from the four above (exact strings), ordered best-fitting first, picking those that match the character of this trajectory. Do not default to a fixed mix.
- If the full content does not actually support a coherent chain toward the given conclusion, output an empty JSON object {{}}.
- Output ONLY the raw JSON object. Do NOT wrap it in markdown formatting or ```json code fences. No explanatory prose.
- Do not refer to "the paper", "the authors", "the study", sections, figures, or tables.

### Paper Content
{paper_content}

### Output Schema
Return ONLY a JSON object with this exact structure:
{{
  "steps": [
    {{
      "step_number": 1,
      "description": "What this step establishes or argues, in clear scientific terms.",
      "observables": ["Key measurement, observation, numerical value, or fact supporting this step"]
    }}
  ],
  "recommended_perspectives": [
    "mechanistic_reasoning",
    "hypothesis_falsification"
  ]
}}
\end{lstlisting}
\end{promptbox}
\caption{Prompt for Claim-Conditioned Trajectory Extraction.}
\label{fig:prompt_trajectory_extraction}
\end{figure*}

\begin{figure*}[t]
\begin{promptbox}{\textit{Perspective Guidance for Question Generation}}
\begin{lstlisting}[style=prompt]
[mechanistic_reasoning]
Target: Probe the reasoning link between a condition, intervention, or structure and an observed phenomenon. Frame this link as a universal scientific problem, abstracting away paper-specific identifiers so that solving it requires general mechanistic reasoning rather than reading comprehension.
Style: Answerable in 2-5 paragraphs of mechanistic reasoning without extensive math. Focus entirely on the system/phenomenon itself.
Construction:
- State the setup using generic descriptors (e.g. "a two-component system under periodic forcing").
- The expected answer should be a chain of mechanistic steps, each justified by a general principle, not a single recalled fact.
Perspective Guardrails:
- Anti-Stipulation: Do NOT stipulate a counterintuitive finding and ask "why". The solver must engage via general principles to deduce the outcome.
- Anti-Comprehension: Must not be a mere reading-comprehension question. If it requires having read the specific paper to start reasoning, reject it.
- Single-Mechanism Focus: Center one dominant causal mechanism. Do not bundle several loosely related effects into one prompt.

[hypothesis_falsification]
Target: Test whether the evidence rules out plausible but competing explanatory models, forcing the solver to identify which single model the evidence is consistent with.
Style: MCQ format. The `question` string MUST end with four options labeled (A), (B), (C), (D) separated by newlines. Do NOT indicate the correct answer in the question.
Construction:
- Present a concrete body of evidence in the stem, then offer four candidate explanations.
- Each distractor must be defeated by a DIFFERENT, identifiable piece of the evidence, so elimination requires genuine discrimination rather than plausibility ranking.
Perspective Guardrails:
- Structural Rivalry: Options must be structurally different explanatory models or historical misconceptions, NOT paraphrases or lexical variants of the same answer.
- Lexical Decoupling: Avoid technical terms in the stem that act as "giveaways" via direct vocabulary matching with the correct option. The correct option must not be identifiable by surface word overlap alone.
- Balanced Distractors: Keep options comparable in length, specificity, and technical register so none stands out by form rather than content.

[quantitative_derivation]
Target: Convert numerical observables into a modeling-first problem. The solver must formulate the relevant physical, chemical, or statistical relations and derive an analytical expression, bound, scaling relation, or key calculated quantity.
Style: A rigorous graduate-level theoretical problem. State knowns objectively and demand derivation.
Construction:
- Provide only the observables genuinely needed; let the solver decide which principles couple them.
- Ask for a symbolic result first (expression, bound, or scaling), with any numerical evaluation as a final, minor step.
Perspective Guardrails:
- Modeling-First: The intended difficulty MUST lie in constructing the model and connecting principles, NOT in tedious arithmetic or plug-and-chug substitution.
- Anti-Fragmentation: Focus on ONE core modeling challenge. Avoid sub-questions that are purely arithmetic continuations.
- No Over-Specification: Do NOT pre-state the governing equation, the regime, or the solution method. Handing the solver the premise removes the modeling step that defines this perspective.

[boundary_calibration]
Target: Target the qualifiers, assumptions, parameter regimes, or confounders under which the claim would no longer be justified.
Style: Concise critique targeting a specific assumption/approximation breakdown (e.g., "At what parameter regime does this approximation break down?").
Construction:
- Anchor on one specific claim, then ask where its supporting evidence stops holding.
- Frame the answer around a concrete switch: a parameter crossing a threshold, an assumption being violated, or a confounder becoming non-negligible.
Perspective Guardrails:
- Sharp Critique: Point to a SPECIFIC, scientifically reachable approximation, regime, or confounder. Do not ask vague "what could go wrong" questions.
- No Sci-Fi: The boundary must be physically or operationally reachable, not an absurd hypothetical (e.g., halving gravity).
- Justified-vs-Unjustified: The item should hinge on distinguishing what the evidence does support from what it does not, not on listing every conceivable caveat.
\end{lstlisting}
\end{promptbox}
\caption{Perspective Guidance for Question Generation.}
\label{fig:prompt_perspective_guidance}
\end{figure*}

\begin{figure*}[t]
\begin{promptbox}{\textit{Shared Prompt for Multi-Perspective Question Generation}}
\begin{lstlisting}[style=prompt]
### Role & Objective
You are an expert data architect crafting high-quality Supervised Fine-Tuning data to train next-generation reasoning models.
Distill a PhD-level scientific problem from the 'Paper Trajectory' based on the requested perspective. The item must function as a rigorous logic stress test requiring complex, multi-step deduction.

### Input Context
- Requested Perspective: {requested_perspective}

### Perspective Focus
{perspective_guidance}

### Global Guardrails & Structural Bans (CRITICAL)
To prevent dataset mode collapse, enforce these rules strictly:
1. Universal Abstraction (No Source Referencing): The problem MUST be entirely self-contained. NEVER use labels like "In this study," "the authors," "the trajectory," or result-reporting verbs ("found," "measured").
2. No Hyperlocal Naming: Replace sample-specific IDs, specific instrument names, or isolated regional names with generic scientific descriptors.
3. No Meta/Lazy Evaluations: NEVER end with meta-questions (e.g., "Is this justified?", "What remains unknown?").
4. Target Difficulty: Difficulty must stem from intricate mechanistic chains, subtle tradeoffs, or advanced modeling rather than niche trivia.

### Global Skip Conditions
Output EXACTLY `[]` (an empty array) with no other text if ANY of the following apply:
- The trajectory lacks the fundamental depth, alternatives, or theoretical bounds required for the `{requested_perspective}`.
- The core conclusion is just a descriptive observation/high-level summary with no "black box" left to interrogate.
- The question cannot be made self-contained without relying on paper-specific data.
- The question degrades into a trivial textbook variant of a standard field problem.

### Output Schema
[
  {{
    "question": "The fully formulated scientific problem (append (A)-(D) options here if MCQ).",
    "evidence": ["Concise scientific fact 1 from trajectory", "Fact 2"],
    "solvability_justification": "Rigorous proof the question is well-posed. For open math: explain theoretical steps. For MCQs: state correct option AND detailed elimination logic for distractors."
  }}
]

### Paper Trajectory
{trajectory_text}
\end{lstlisting}
\end{promptbox}
\caption{Shared Prompt for Multi-Perspective Question Generation.}
\label{fig:prompt_question_generation}
\end{figure*}

\begin{figure*}[t]
\begin{promptbox}{\textit{Prompt for Answer Generation}}
\begin{lstlisting}[style=prompt]
{question}
Solve the above problem step by step.
\end{lstlisting}
\end{promptbox}
\caption{Prompt for Answer Generation.}
\label{fig:prompt_answer_generation}
\end{figure*}

\begin{figure*}[t]
\begin{promptbox}{\textit{Prompt for Skeleton-Grounded Quality Checking}}
\begin{lstlisting}[style=prompt]
### Role & Objective
You are a scientific QA auditor. Rule-based checks already catch format failures; your job is the part rules cannot decide: whether this QA pair is scientifically grounded in the paper trajectory. Judge it for reasoning-model SFT training and return a single KEEP/DISCARD decision.

### What You Receive
- A Paper Trajectory: the extracted core conclusion (the scientific direction) plus the step-by-step evidence chain (the evidence anchors).
- A Generated Question and a Generated Answer to audit against that trajectory.

### Audit Dimensions

#### Dimension 1: Question
1. Self-Containment
The question must provide the conditions, variables, and context needed to be answered WITHOUT the original paper. All custom variables, experimental settings, and boundary conditions must be stated. Common universal scientific constants need not be restated. Flag any question that silently depends on unstated paper-specific content.

2. Source Match
The question must correspond to what the trajectory actually establishes. Flag a source mismatch when the question concerns an entity, quantity, or claim the trajectory does not support, or fabricates premises absent from it.

#### Dimension 2: Answer
1. Reasoning Coherence
The reasoning chain must be coherent and complete: no logical jumps, no meaningless repetition or loops, no interrupted or truncated output. The final answer must be definite and well-organized.
*Rigorous self-correction, re-evaluation of earlier steps, and explicitly resolving uncertainty within the text are valuable and fully permitted.*

2. Trajectory Grounding
The answer must be consistent with the scientific direction of the trajectory's core conclusion AND not be contradicted by its evidence chain (steps and observables). Equivalent academic statements, synonymous expressions, and reasonable paraphrased reasoning are acceptable; only essential conflicts with the established direction or evidence are disqualifying.

### Hard Elimination Rules
1. Dependency Defect: Missing essential custom variables; unanswerable without extra paper context. -> DISCARD
2. Source Mismatch: Question targets claims/entities the trajectory does not support, or invents absent premises. -> DISCARD
3. Generation Defect: Incoherent reasoning -> meaningless loops, arbitrary guessing without deduction, truncated output, or serious hallucination. -> DISCARD
4. Grounding Conflict: Answer's core direction is inconsistent with the trajectory's conclusion, or is contradicted by its evidence chain. -> DISCARD

Supplementary Rule: Standard multi-step inference, causal analysis, and logical self-correction all count as qualified responses.

### Output Specification
Output a single, strictly valid JSON object. Do NOT add any extra symbols, markdown formatting outside the JSON, or conversational text. Use the exact schema below:

{{
  "reasoning_steps": "First, write your thorough step-by-step reasoning and analysis here.",
  "checks": {{
    "is_question_self_contained": true,
    "is_question_source_matched": true,
    "is_answer_coherent": true,
    "is_answer_trajectory_grounded": true
  }},
  "fatal_flaw_details": "Concrete flaw reason based on the Hard Elimination Rules, or 'none'.",
  "final_decision": "KEEP"
}}
// Note: "final_decision" must be strictly either "KEEP" or "DISCARD".

### Paper Trajectory
{trajectory_text}

### Generated Question
{question}

### Generated Answer
{answer}
\end{lstlisting}
\end{promptbox}
\caption{Prompt for Skeleton-Grounded Quality Checking.}
\label{fig:prompt_quality_check}
\end{figure*}

%==============================================================================
\section{Source Leakage Patterns}
\label{sec:appendix:leakage}
%==============================================================================

We apply a deterministic source-leakage filter to remove questions that
implicitly require access to the original document. The filter consists of 21
case-insensitive regular-expression patterns and is applied to generated
questions before they enter the final dataset. Table~\ref{tab:source_leakage_patterns}
summarizes the pattern groups. Rather than attempting to judge all forms of
self-containment, this filter targets explicit linguistic cues that a question
is referring back to the source paper, provided context, or document structure.
Any record matching one or more patterns is discarded.

\begin{table}[H]
  \centering\small\setlength{\tabcolsep}{4pt}
  \caption{Summary of source-leakage pattern groups.}
  \label{tab:source_leakage_patterns}
  \begin{tabular}{p{2.9cm}p{4.5cm}}
    \toprule
    \textbf{Pattern group} & \textbf{Representative cues} \\
    \midrule
    Source-document references &
    ``the paper'', ``the study'', ``the authors'', ``the article'', ``the passage'', ``this research'' \\
    Provided-context references &
    ``provided text'', ``provided material'', ``provided information'', ``provided context'', ``according to the provided'' \\
    Internal evidence references &
    ``supporting evidence'', ``supporting facts'', ``scientific context'', ``internal context'', ``internal scientific facts'' \\
    Text-reporting constructions &
    ``the text states'', ``the text shows'', ``the text reports'', ``the text demonstrates'', and related verb variants \\
    Document-element references &
    ``section $N$'', ``figure $N$'', ``table $N$'', ``reference $N$'', ``appendix $N$'' \\
    \bottomrule
  \end{tabular}
\end{table}

%==============================================================================
\section{Difficulty Taxonomy}
\label{sec:appendix:difficulty}
%==============================================================================

We evaluate question difficulty with a five-level reasoning taxonomy. For each
dataset, we sample 20K questions and ask DeepSeek-V4-Flash to assign one of the
following levels according to the minimum reasoning required to obtain the
answer:

\begin{itemize}
    \item \textbf{L1 --- Fact Retrieval.} The answer can be directly retrieved
    as a fact, definition, or given value from knowledge or the source. No
    reasoning, derivation, or inference is required.
    \item \textbf{L2 --- Routine Calculation.} The answer is obtained by
    applying a known formula or standard procedure and substituting the provided
    values. No modeling, method selection, or multi-step reasoning is required.
    \item \textbf{L3 --- Single-Step Reasoning.} The question requires one
    non-trivial inference, conceptual link, or causal/mechanistic explanation
    beyond direct retrieval or mechanical substitution.
    \item \textbf{L4 --- Multi-Step Reasoning.} The question requires a chain of
    reasoning steps, mechanistic analysis, or discrimination among multiple
    competing explanations or hypotheses.
    \item \textbf{L5 --- Research-Level Reasoning.} The question requires
    constructing a model from principles, synthesizing multiple pieces of
    evidence, carrying out a derivation, or analyzing boundary conditions,
    approximations, or counterfactuals.
\end{itemize}

The resulting labels are used only for aggregate analysis rather than as a
training signal. This protocol is intended to compare the reasoning profile of
datasets under a consistent rubric, not to provide definitive expert judgments
for individual examples.

\section{Human Validation of Quality Assessments}
\label{sec:appendix:human_validation}

To validate the reliability of our LLM-based quality assessments, we conduct a stratified human audit of 300 samples, with 30 questions drawn from each of the ten disciplines. Two paid domain experts for each discipline independently evaluate the samples while remaining blind to the original answers and model-assigned labels:

\begin{itemize}
\item \textbf{Correctness:} Experts independently solve each question, and their adjudicated answer is compared with the original answer in the dataset.
\item \textbf{Self-containment:} Experts judge whether the question contains all information needed to be answered independently.
\item \textbf{Difficulty:} Experts assign one of the five reasoning levels (L1--L5) defined in Appendix~\ref{sec:appendix:difficulty}.
\end{itemize}

Disagreements are resolved through discussion. Table~\ref{tab:human_validation} reports three metrics for each dimension: \textit{Human Assessment} (HA) reports the proportions judged self-contained and assigned L4/L5; it is not applicable to correctness because correctness is evaluated through open-ended problem solving. \textit{Agreement with Original Dataset Result} (Agreement) compares the adjudicated human labels with the original LLM-assigned labels (or the independently derived expert answers with the original answers for correctness). \textit{Inter-Annotator Agreement} (IAA) is computed before adjudication.

\begin{table}[H]
\centering
\caption{Human validation results for correctness, self-containment, and difficulty.}
\label{tab:human_validation}
\resizebox{\columnwidth}{!}{%
\begin{tabular}{lccc}
\toprule
\textbf{Dimension} & \textbf{HA} & \textbf{Agreement} & \textbf{IAA} \\
\midrule
Correctness & NA & 90.3\% (271/300) & 96.0\% (288/300) \\
Self-containment & 98.0\% (294/300) & 97.7\% (293/300) & 99.3\% (298/300) \\
Difficulty (L4/L5) & 87.0\% (261/300) & 94.3\% (283/300) & 91.0\% (273/300) \\
\bottomrule
\end{tabular}%
}
\end{table}

The 90.3\% agreement with independently derived expert answers supports the reliability of the original answers. Human experts also judge 98.0\% of the questions to be self-contained and 87.0\% to be L4/L5. Their agreement with the original LLM-assigned labels reaches 97.7\% and 94.3\%, respectively, supporting the reliability of the automatic assessments. Difficulty is inherently more subjective, as also reflected by its lower inter-annotator agreement; nevertheless, the human results confirm that the dataset predominantly contains challenging questions.

%==============================================================================
\section{Diversity Metric Calculations}
\label{sec:appendix:diversity_metrics}
%==============================================================================

To rigorously evaluate the semantic span of \oursdata{} and baseline data, we
use two complementary metrics: Vendi Score~\citep{friedman2022vendi}
and Centroid Distance~\citep{suwanda2020analysis}. Let
$X=\{x_1,x_2,\dots,x_N\}$ be the set of embeddings for the instructions or
questions in a dataset, where $N$ is the sample size. All embeddings are
computed with \texttt{Qwen3-Embedding-8B}~\citep{zhang2025qwen3}, and the
embedding dimensionality is fixed to 4096 for all samples. Unless otherwise
specified, distance is measured by cosine distance,
$d(x_i,x_j)=1-\frac{x_i\cdot x_j}{\|x_i\|\|x_j\|}$.

\paragraph{Vendi Score.}
The Vendi Score measures intrinsic diversity by interpreting a dataset's
semantic spread as the effective number of independent modes. It is calculated
from the eigenvalues of a kernel matrix $K$, where $K_{ij}=k(x_i,x_j)$ denotes
the similarity between samples. We use the cosine-similarity kernel. The Vendi
Score is defined as the exponential of the Shannon entropy of the normalized
eigenvalues:
\begin{equation}
    \mathrm{Vendi}(X)
    =
    \exp\left(
    -\sum_{i=1}^{N}\lambda_i \ln \lambda_i
    \right),
\end{equation}
where $\lambda_1,\dots,\lambda_N$ are the normalized eigenvalues of $K/N$. A
higher Vendi Score indicates that the dataset contains a larger number of
effective independent semantic clusters. In practice, we compute Vendi Score
using the reference implementation released by the original authors.\footnote{\url{https://github.com/vertaix/Vendi-Score}}

\paragraph{Centroid Distance.}
Centroid Distance measures geometric dispersion in the embedding space. We first
compute the global centroid $\mu$ of the dataset:
\begin{equation}
    \mu = \frac{1}{N}\sum_{i=1}^{N}x_i .
\end{equation}
Centroid Distance is then defined as the complement of the average cosine
similarity between each sample $x_i$ and the centroid $\mu$:
\begin{equation}
    \mathrm{Dist}_{\mathrm{cent}}(X)
    =
    1-\frac{1}{N}\sum_{i=1}^{N}
    \frac{x_i\cdot\mu}{\|x_i\|\|\mu\|}.
\end{equation}
A higher Centroid Distance indicates that samples are more widely dispersed
around the dataset center, covering a broader semantic region rather than
clustering tightly around a narrow set of topics.

%==============================================================================
\section{Self-Containment Bad Cases}
\label{sec:appendix:self_containment_bad_cases}
%==============================================================================

Figure~\ref{fig:self_containment_bad_cases} shows representative self-containment
failures observed in baseline data. These examples illustrate why a fluent
question can still be unusable for standalone reasoning: it may refer to missing
definitions, omitted equations, or unstated variables.

\begin{figure*}[t]
\begin{databox}{\textit{Self-Containment Bad Cases from Baseline Data}}
\small
\textbf{MegaScience.}

\textbf{Case 1: Missing definitions.}
\begin{quote}
Show that if a binary sequence $\langle X_n \rangle$ is R5-random (as per Definition R5), and if $\langle s_n \rangle$ is any computable sequence as in Definition R4, then $\overline{\Pr}(X_{s_n}=1) \geq \frac{1}{2}$ and $\Pr(X_{s_n}=1) \leq \frac{1}{2}$.
\end{quote}
\textbf{Reason:} The question refers to Definition R5 and Definition R4, but the definitions are not provided. A solver cannot independently interpret the assumptions.

\medskip
\textbf{Case 2: Missing referenced equation.}
\begin{quote}
Prove why the last equality holds true for a function in the Dirichlet space.
\end{quote}
\textbf{Reason:} The phrase ``the last equality'' refers to an omitted equation, so the target statement to be proved is unavailable.

\medskip
\textbf{Case 3: Missing required variable.}
\begin{quote}
A roller coaster is moving at an initial velocity of 10 m/s and comes to a stop after a distance of 50 m. Calculate the average force supplied by the brakes to stop the roller coaster. Assume a constant acceleration and neglect any frictional forces. Use the following equations: $V_1=10$ m/s, $\Delta D=50$ m, $V_2=0$ m/s, $F_{\mathrm{net}}=F_{\mathrm{brakes}}=ma$.
\end{quote}
\textbf{Reason:} The mass of the roller coaster is not specified, so the average braking force cannot be determined.

\medskip
\textbf{SCP-116K.}

\textbf{Case 4: Missing circuit diagram.}
\begin{quote}
Two AC signals, $V_1$ and $V_2$, are to be combined such that
$V_{\mathrm{out}}=\frac{3}{2}V_2-\frac{5}{2}V_1$. The subtracting amplifier circuit shown is used. What must be the values of $R_1$, $R_2$, $R_3$, and $R_4$? Circuit diagram: $V_1$ connected via $R_1$; $V_2$ connected via $R_2$; operational amplifier with feedback resistance $R_3$; ground connection via $R_4$. (Professional Publications, Inc.)
\end{quote}
\textbf{Reason:} The question refers to an external circuit diagram. Without the actual diagram, the resistor topology and value relationships are under-specified.

\medskip
\textbf{Case 5: Missing section dependency.}
\begin{quote}
Using the results of Section 7.5, find the transformation $x=Tw$ that puts the system $\frac{dx}{dt}=\begin{pmatrix}1&-1&2\\0&-2&1\\-1&3&0\end{pmatrix}x+\begin{pmatrix}1&3\\0&1\\0&2\end{pmatrix}u$ into the form $\frac{dw}{dt}=\begin{pmatrix}0&-3&6\\-1&-1&0\\0&1&0\end{pmatrix}w+\begin{pmatrix}1&0\\0&0\\0&1\end{pmatrix}u$.
\end{quote}
\textbf{Reason:} The question depends on results from Section 7.5, but those results are not provided. A solver cannot recover the required transformation rule from the question alone.

\medskip
\textbf{Case 6: Missing referenced formulas.}
\begin{quote}
Calculate the Hilbert transforms of $\cos(\omega_0 t)$ and $\sin(\omega_0 t)$. Hint: Use Euler's identity in the defining integral (7.10.12) for the Hilbert transform of a time signal, and you may also find that recalling (7.9.3) is helpful.
\end{quote}
\textbf{Reason:} The prompt relies on formula numbers (7.10.12) and (7.9.3), but the formulas themselves are omitted.

\medskip
\textbf{NaturalReasoning.}

\textbf{Case 7: Undefined property.}
\begin{quote}
What should be the cardinality of $A_n$ to ensure that the subset of infinite binary sequences having a certain property is countable/uncountable, considering the set $R_n$ of binary sequences of length $n$ and the subset $A_n$ of sequences having that property?
\end{quote}
\textbf{Reason:} The question never defines ``a certain property'', so the solver cannot determine which sequences belong to $A_n$ or what condition is being analyzed.

\medskip
\textbf{Case 8: Missing humidity information.}
\begin{quote}
Given the temperature of 18 degrees Celsius and the atmospheric pressure of 782 mm Hg, calculate the air density using the ideal gas law equation, considering the effects of humidity. Show your work and explain your reasoning.
\end{quote}
\textbf{Reason:} The question asks the solver to account for humidity but provides no relative humidity, water-vapor partial pressure, or equivalent humidity-related quantity.
\end{databox}
\caption{Representative self-containment failures from baseline data. Each example appears plausible but lacks information required for standalone solution.}
\label{fig:self_containment_bad_cases}
\end{figure*}

%==============================================================================
\section{Qualitative Examples}
\label{sec:appendix:examples}
%==============================================================================

To provide qualitative insight into the output of our automated data construction pipeline, we present examples of a generated paper skeleton and corresponding generated questions and answers in Figure~\ref{fig:synthesis_pipeline}. The figure illustrates the clear structure extracted for a sample paper, highlighting key entities, processes, and findings. Linked to specific points in this skeleton are sequential examples of different question types—entity-based, procedural, causal, and more—each accompanied by a grounded answer derived directly from the skeleton content. Subtle visual links demonstrate the grounding and diversity of the generated data, ensuring a robust and factually anchored fine-tuning dataset for scientific reasoning.

\begin{figure*}[t]
\begin{databox}{\textit{Dataset Synthesis: From Structured Extraction to Deep Reasoning QA}}
\small
\textbf{Phase 1: Extracted Paper Framework}

\textit{Core Conclusion:} A tail wave formed behind density cusps in nonlinear laser wakefields driven by a Gaussian laser pulse provides simultaneous focusing and acceleration for externally injected positrons, enabling nearly 100\% trapping efficiency with terawatt-class laser systems.

\begin{itemize}[leftmargin=*, parsep=0pt, itemsep=6pt, topsep=4pt]
    \item \textbf{Step 1: Defocusing mechanism \& Tail wave proposal.}
    \begin{itemize}[leftmargin=1.5em, parsep=0pt, itemsep=2pt, topsep=2pt]
        \item \textit{Description:} Establish that nonlinear wakefields naturally defocus positively charged particles. Propose that a tail wave forms behind density cusps when laser intensity marginally exceeds the nonlinear threshold, converging plasma electrons into an on-axis filament.
        \item \textit{Observables:} $a_0 = 1.6$, $w_0 = 12\lambda_0$, peak intensity $3.5\times10^{18}$ W/cm$^2$. Regime thresholds based on $n(0)/n_{e0}$: I ($\geq0.2$), II ($0.1\text{--}0.2$), III ($\leq0.1$).
    \end{itemize}

    \item \textbf{Step 2: Unloaded dynamics.}
    \begin{itemize}[leftmargin=1.5em, parsep=0pt, itemsep=2pt, topsep=2pt]
        \item \textit{Description:} Using 3D PIC simulations (OSIRIS), demonstrate tail wave dynamics without positron loading. The focusing region for positrons expands from a limited cusp to about half a bubble as laser intensity increases.
        \item \textit{Observables:} Grid $200\times200\times2400$. Transverse field maps show focusing region expanding.
    \end{itemize}

    \item \textbf{Step 3: Beam loading \& Field reversal.}
    \begin{itemize}[leftmargin=1.5em, parsep=0pt, itemsep=2pt, topsep=2pt]
        \item \textit{Description:} Load a uniform positron beam into the second bubble. The beam's space-charge attracts tail wave electrons, converting a defocusing transverse field into a focusing field. However, tail wave III causes transverse loss.
        \item \textit{Observables:} Positron $n_{+0} = 0.5n_{e0}$, initial energy 20 MeV. Trapping efficiency $\sim60\%$ due to tail wave III.
    \end{itemize}

    \item \textbf{Step 4: Channel optimization.}
    \begin{itemize}[leftmargin=1.5em, parsep=0pt, itemsep=2pt, topsep=2pt]
        \item \textit{Description:} Increase the plasma channel radius to weaken relativistic self-focusing and suppress tail wave III, maintaining the required on-axis electron density throughout propagation.
        \item \textit{Observables:} Channel radius $r_0 = 14\lambda_0 > w_0$. Density $n(0)/n_{e0} \geq 0.1$ (avoiding regime III). Trapping efficiency $>95\%$ after 1000$\lambda_0$.
    \end{itemize}

    \item \textbf{Step 5: Beam evolution.}
    \begin{itemize}[leftmargin=1.5em, parsep=0pt, itemsep=2pt, topsep=2pt]
        \item \textit{Description:} Characterize the accelerated positron beam, showing reduced transverse size, limited emittance growth, and stable longitudinal acceleration.
        \item \textit{Observables:} Transverse size reduced to $\sim1/3$. Acceleration gradient $\sim100$ GV/m. Final energy peak at 177 MeV (average gain 110 MeV) with 3\% FWHM energy spread.
    \end{itemize}

    \item \textbf{Step 6: Parameter robustness.}
    \begin{itemize}[leftmargin=1.5em, parsep=0pt, itemsep=2pt, topsep=2pt]
        \item \textit{Description:} Vary drive laser parameters and beam offsets to map the viability of the scheme.
        \item \textit{Observables:} $a_0 \in [1.2, 1.8]$, $w_0 \in [8\lambda_0, 12\lambda_0]$ for high trapping. Longitudinal offset tolerance within $2\lambda_0$; transverse tolerance $\sim0.5\lambda_0$.
    \end{itemize}
\end{itemize}

\vspace{0.5em}
\hrule
\vspace{0.5em}

\textbf{Phase 2: Synthesized SPARK-QA}

\begin{quote}
In a laser-driven plasma wakefield used for positron acceleration, the transverse field that a witness positron beam will experience is predicted from a simulation of the \emph{bare} wake---one run without the beam present (an ``unloaded'' field map). A positron beam of density $n_+ = 0.5\,n_{e0}$ ($n_{e0}$ = ambient plasma density) is then injected into an accelerating bucket of this wake, and its focusing or defocusing is read off from the sign of the unloaded transverse field at the beam's location. Which of the following is the most fundamental limitation of using the unloaded field map to predict the focusing experienced by the \emph{loaded} beam?

\begin{enumerate}[label=(\Alph*)]
  \item As the drive laser propagates, relativistic self-focusing reshapes the
        pulse and therefore the wake's longitudinal profile, so a field map
        taken at one propagation distance no longer applies further downstream.
  \item The guiding plasma channel imposes a transverse density gradient that
        the bare-wake simulation does not reproduce, so the off-axis transverse
        field is mis-estimated across the beam's radial extent.
  \item The positron beam's own space charge draws plasma electrons toward the
        axis, adding on-axis negative charge that the unloaded map omits; this
        can convert a region that is defocusing for positrons in the bare wake
        into a focusing one.
  \item The transverse field varies strongly with longitudinal phase within a
        single bucket, so a map read at one slice of the bucket mis-predicts
        the field at the precise phase the beam occupies.
\end{enumerate}
\end{quote}
\end{databox}
\caption{Dataset synthesis pipeline. Phase 1 details the extraction of a logically closed paper framework, structurally decoupling qualitative descriptions from quantitative observables. Phase 2 leverages this structured context to generate a self-contained, expert-level scientific reasoning question.}
\label{fig:synthesis_pipeline}
\end{figure*}

\section{Use of AI Assistants}
We used a large language model only for light writing assistance, limited to grammar checking and minor clarity edits on a small number of sentences.
All technical content, experiments, and analyses in this paper were written by the authors.

\end{document}